\documentclass[10 pt, journal, twoside]{ieeetran}
\usepackage{amsmath,epsfig}
\usepackage{setspace}
\usepackage{bm}
\usepackage{cite}
\usepackage{color}
\usepackage{float} 
\usepackage{array}
\usepackage{bbding}
\usepackage{pifont}
\usepackage{amssymb}
\usepackage{wasysym}
\usepackage{pdfpages}
\usepackage{epstopdf}
\usepackage{verbatim}
\usepackage{threeparttable}
\usepackage[colorlinks=true, linkcolor=blue, anchorcolor=blue, citecolor=blue]{hyperref}
\usepackage{lipsum,cite,multicol,multirow,amssymb,graphicx,array,epstopdf,subfigure,makecell,color, algorithm, algorithmic,booktabs}

\ifCLASSINFOpdf
\else
\usepackage[dvips]{graphicx}
\fi
\usepackage{url}
\usepackage{graphicx}

\newcommand{\PreserveBackslash}[1]{\let\temp=\\#1\let\\=\temp}

\newcolumntype{C}[1]{>{\PreserveBackslash\centering}p{#1}}
\newcolumntype{R}[1]{>{\PreserveBackslash\raggedleft}p{#1}}
\newcolumntype{L}[1]{>{\PreserveBackslash\raggedright}p{#1}}
\newcommand{\etal}{\textit{et al}.}
\newcommand{\ie}{\textit{i}.\textit{e}.}
\newcommand{\eg}{\textit{e}.\textit{g}.}

\begin{document}
	\title{UIERL: Internal-External Representation Learning Network for Underwater Image Enhancement}
	\author{Zhengyong Wang, Liquan Shen, Yihan Yu and Yuan Hui}
\maketitle
\begin{abstract} 
Underwater image enhancement (UIE) is a meaningful but challenging task, and many learning-based UIE methods have been proposed in recent years. Although much progress has been made, these methods still exist two issues: (1) There exists a significant region-wise quality difference in a single underwater image due to the underwater imaging process, especially in regions with different scene depths. However, existing methods neglect this internal characteristic of underwater images, resulting in inferior performance; (2) Due to the uniqueness of the acquisition approach, underwater image acquisition tools usually capture multiple images in the same or similar scenes. Thus, the underwater images to be enhanced in practical usage are highly correlated. However, when processing a single image, existing methods do not consider the rich external information provided by the related images. There is still room for improvement in their performance. Motivated by these two aspects, we propose a novel internal-external representation learning (UIERL) network to better perform UIE tasks with internal and external information, simultaneously. In the internal representation learning stage, a new depth-based region feature guidance network is designed, including a region segmentation based on scene depth to sense regions with different quality levels, followed by a region-wise space encoder module. With performing region-wise feature learning for regions with different quality separately, the network provides an effective guidance for global features and thus guides intra-image differentiated enhancement. In the external representation learning stage, we first propose an external information extraction network to mine the rich external information in the related images. Then, internal and external features interact with each other via the proposed external-assist-internal module (external features are updated with the help of internal features) and internal-assist-external module (internal features are updated with the help of external features). In this way, our UIERL fully explores the rich internal and external information to better enhance a single image. All results show that our method can achieve state-of-the-art performance on five benchmarks.
\end{abstract}

\section{Introduction}
\begin{figure}[!t]
	\centering
	\centerline{\includegraphics[height=5.9cm,width=8.92cm]{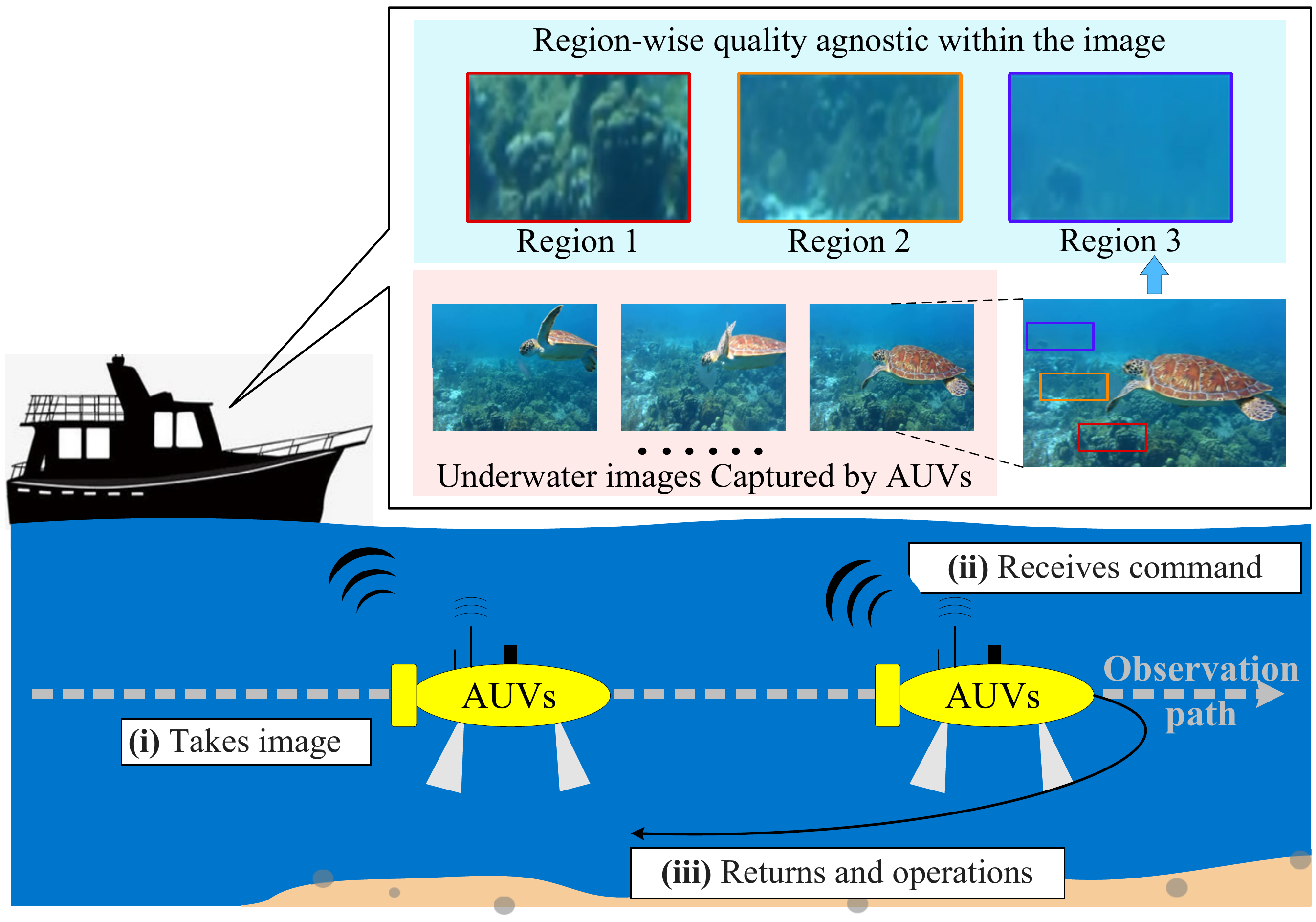}}
	\caption{\textbf{Motivations of our proposed UIERL}. (1) Due to the unique underwater imaging process, there exists a significant region-wise quality difference in a single underwater image, especially in regions with different scene depth values, \ie, region-wise quality agnostic within the image. (2) Due to the uniqueness of the acquisition approach, underwater image acquisition tools (AUVs or ROVs) usually capture multiple images in the same or similar scenes. Thus, the underwater images to be enhanced in practical applications are highly correlated. These related images can provide rich complementary information for each other to further improve performance.}
	\label{Fig1}
\end{figure}

\IEEEPARstart{B}ecause of the substantial resources presented in natural and biological cases, underwater observation and exploration have attracted increasing attention in the last few years. Unlike diverse observation and exploration approaches on land, researchers cannot dive into a deep position for observation or engineering due to the limitations of the marine environment. Thus, humans usually resort to autonomous underwater vehicles (AUVs) and remotely operated vehicles (ROVs) to explore underwater space and photograph underwater images on the pre-determined observation path, as shown in Fig.\ref{Fig1}. However, the quality of captured underwater images is always unsatisfactory. They often suffer from severe distortions such as color casts, low contrast and blurred details due to light absorption and scattering. This imposes many limitations on subsequent visual perception analysis and exploration of the underwater world. There is an urgent need to improve the quality of underwater images. 

During the past few years, rapidly developed deep learning techniques~\cite{Song2018, Yang2019, Jin2020, Wang2021} have further propelled the advancement of learning-based underwater image enhancement (UIE) approaches. Numerous deep methods specially designed for underwater images have been proposed in this community. These methods have made considerable efforts on designing new sample generation technologies~\cite{li2017watergan, yu2018underwater, li2019underwater, Liu2019, Li2020, dudhane2020deep, 2021Single}, more effective learning strategies and network architectures~\cite{fabbri2018enhancing, li2018emerging, guo2019underwater, Liu2020, Lin9311705, wang9392321, Xue2021, Jiang2022a, Wang2023, Jiang2022}, or leveraging some prior knowledge~\cite{islam2020fast, Li2021, Shi2021, Fu2022, Li2022, Qi2022} to mine as much useful information as possible from a single image. Although remarkable progress has been achieved in this field, there are still two issues limiting their performance.
\begin{figure*}[t]
	\centering
	\centerline{\includegraphics[width=18cm, height=7cm]{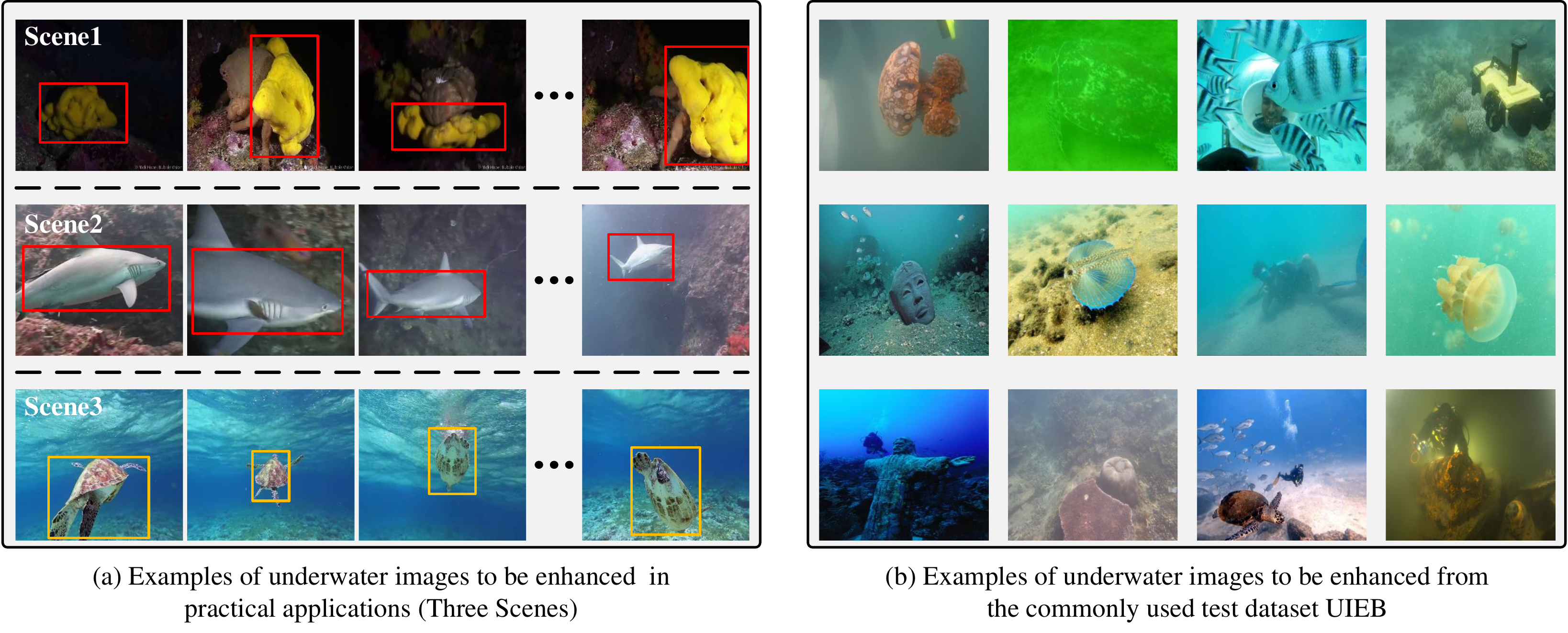}}
	\caption{Comparison of underwater images to be enhanced captured by AUVs and ROVs with the commonly used test dataset UIEB~\cite{li2019underwater}. As shown, the UIEB dataset does not fully reflect the characteristics of underwater images captured in the real world. It can clearly be seen that these underwater images to be enhanced in practical applications are highly correlated. They can provide rich complementary information for each other’s enhancement.}
	\label{Fig2}
\end{figure*}

First, there exists a significant region-wise quality difference within a single underwater image due to the unique underwater imaging process, particularly referring to regions with different scene depth values. Unlike terrestrial images where light absorption is assumed to be spectrally uniform and light scattering is neglected, multiple underwater distortions cannot be corrected globally since the light absorption and scattering often vary in the water body and scene depth~\cite{chiang2011underwater, Akkaynak2017, Akkaynak2018, Akkaynak2019}. This varying characteristic of underwater light results in an obvious region-wise quality agnostic within the image, as shown in the first row of Fig.\ref{Fig1}. Ideally, for regions with different quality levels, the enhancement network should impart different degrees of response, \ie, differentiated enhancement. However, existing methods neglect the difference in the quality of different regions within one single image; thus, they process the whole image in the same manner.

Second, there exists a wealth of rich external information that can be used to help enhance a single underwater image. Due to the uniqueness of the acquisition approach, underwater image acquisition tools (AUVs or ROVs) usually capture multiple images in the same or similar scenes. Fig.\ref{Fig2} shows the comparison of the underwater images to be enhanced captured by AUVs and ROVs with the commonly used test dataset UIEB~\cite{li2019underwater}. As shown, the UIEB dataset does not fully reflect the characteristics of real underwater test data in practical applications. It is observed that the underwater images to be enhanced in practical usage are highly correlated, and they are usually associated with each other in different ways such as similar color tones, common objects and relevant scenes. These related underwater images can provide rich complementary information for each other to further improve performance. On one hand, multiple images present the same content but with different quality. For example, as shown in the red boxes in Fig.\ref{Fig2} (a), one object may appear much clearer on a close shot region than it on distant views, so the higher quality content can be used to guide the heavily-degraded content enhancement. On the other hand, similar objects might lose and retain diverse information on different images even if all the objects are of low quality as shown in the yellow boxes in Fig.\ref{Fig2} (a), which can be combined to use the complementary information for better content reconstruction. However, when processing a single image, existing methods mainly focus on using single image information, and do not take into account the rich external information provided by the related images. There is still improvement room in their performance. 

Motivated by the above two analyses, this paper proposes a novel \textbf{I}nternal-\textbf{E}xternal \textbf{R}epresentation \textbf{L}earning (UIERL) network for better underwater image enhancement. The proposed UIERL includes two stages: an internal representation learning stage and an external representation learning stage. Concretely, a new depth-based region feature guidance (DRFG) network is designed in the first stage to guide intra-image differentiated enhancement. DRFG consists of a region segmentation based on scene depth map and a region-wise space encoder module, one for perceiving regions with different quality levels and another for learning region-wise features of different quality regions separately. With performing independent learning in different quality regions, the network can obtain rich region features with quality clues, which can be an effective complement and guidance for global-wise learning.

In the second stage, a simple yet efficient external information extraction network (EIEN) is first proposed to capture the rich external information in the related images. Then, internal features and external features interact with each other via the proposed external-assist-internal module (EAI, where external features are updated with the assistance of internal features) and internal-assist-external module (IAE, where internal features are updated with the assistance of external features). UIERL incorporates these two interaction modules which are complementary with each other to fully utilize both internal and external information for UIE. The overview of UIERL is shown in Fig.\ref{Fig3}. To our best knowledge, this is the first work that jointly explores internal and external information in the underwater image enhancement community. The main contributions of this paper are highlighted as follows:

(1) A novel internal-external representation learning network is proposed for UIE tasks, called UIERL, which effectively combines the advantages of both external and internal information, and successfully sheds new light on the future direction of enhancing underwater images.

(2) A new depth-based region feature guidance network is proposed to guide intra-image differentiated enhancement by exploiting the inherent characteristic of underwater degradation, including a region segmentation to perceive regions with different quality levels, and a region-wise space encoder module to build regional cues for global learning.

(3) An external information extraction network is proposed to mine the rich external information in the related images. Besides, we introduce two strategies, \ie, external-assist-internal module and internal-assist-external module, to achieve the full interaction of internal and external features. In this manner, the network combines both rich internal and external information, which can better enhance a single image.

\section{Related Work}
\subsection{Deep Learning-based UIE Methods}
With the great success of deep learning techniques in the field of low-level computer vision, many deep UIE methods have been proposed in recent years. These methods have made extensive efforts on new underwater sample generation techniques~\cite{li2017watergan, yu2018underwater, li2019underwater, Liu2019, Li2020, dudhane2020deep, 2021Single}, more efficient learning strategies and network architectures~\cite{fabbri2018enhancing, li2018emerging, guo2019underwater, Liu2020, Lin9311705, wang9392321, Xue2021, Jiang2022a, Wang2023, Jiang2022}, and well-designed prior information guidance mechanisms~\cite{islam2020fast, Li2021, Shi2021, Fu2022, Li2022, Qi2022}.

Unlike other low-level vision tasks such as deraining and dehazing, it is impractical to obtain both distorted and dewatered images of the same underwater scene. Researchers propose to use synthetic data or pseudo reference data to construct samples for model training. For example, based on the commonly used underwater imaging model, Li \etal~\cite{Li2020} propose the first underwater synthetic dataset, which includes 10 water types. Wang \etal~\cite{2021Single} propose a more comprehensive underwater physical imaging model, which contains an improved physical model that takes more parameter dependencies into account and an unsupervised CNN model works as a supplement to simulate other influencing factors. Thanks to the great success of style transfer techniques, Fabbri \etal~\cite{yu2018underwater} adopt GAN to generate paired underwater training data. Then, GAN-based UIE methods are widely explored, such as FUnIE-GAN~\cite{islam2020fast}, Water-CycleGAN~\cite{li2018emerging}, DenseGAN~\cite{guo2019underwater} and MFFN~\cite{Liu2020}. Later, Li \etal~\cite{li2019underwater} build the first real underwater image enhancement dataset, including 890 paired images. All images are enhanced by 12 UIE algorithms, and manually selected the best results as pseudo references according to subject visual preference. Liu \etal~\cite{liu2019real} propose a large-scale real underwater image enhancement benchmark under natural light, including three subsets for three challenging aspects, \ie, image visibility quality, color casts, and high-level tasks. 

To improve the performance of underwater image enhancement, researchers also explore new learning strategies, network structures and prior information guidance mechanisms. For example, Li \etal~\cite{li2019underwater} propose a gate fusion network to blend three pre-processing images generated by white balance, histogram equalization and gamma correction methods, obtaining a better result. Focusing on solving the color casts and low contrast problems, Li \etal~\cite{Li2021} propose a multi-color space embedding encoder coupled with a new medium transmission-guided decoder. Wang \etal~\cite{Wang2023} consider the underwater image enhancement problem from inter- and intra-domain perspectives, and propose a novel two-phase underwater domain adaptation method. Jiang \etal~\cite{Jiang2022} consider the degradation of underwater images in terms of both turbidity and chromatism, and propose a target oriented perceptual adversarial fusion network for underwater image enhancement. Qi \etal~\cite{Qi2022} introduce semantic information as the high-level guidance to better achieve robustness towards unknown scenarios.

Although these methods have made remarkable progress, they do not fully consider the internal degradation characteristic of underwater images (\ie, region-wise quality agnostic within the image), and thus their performance is severely limited. Furthermore, when processing a single image, these methods only utilize the information within one single image, and neglect the rich external information provided by the related images in practical applications. There is still room for improvement in their performance.
\begin{figure*}[!t]
	\centering
	\centerline{\includegraphics[scale=0.79]{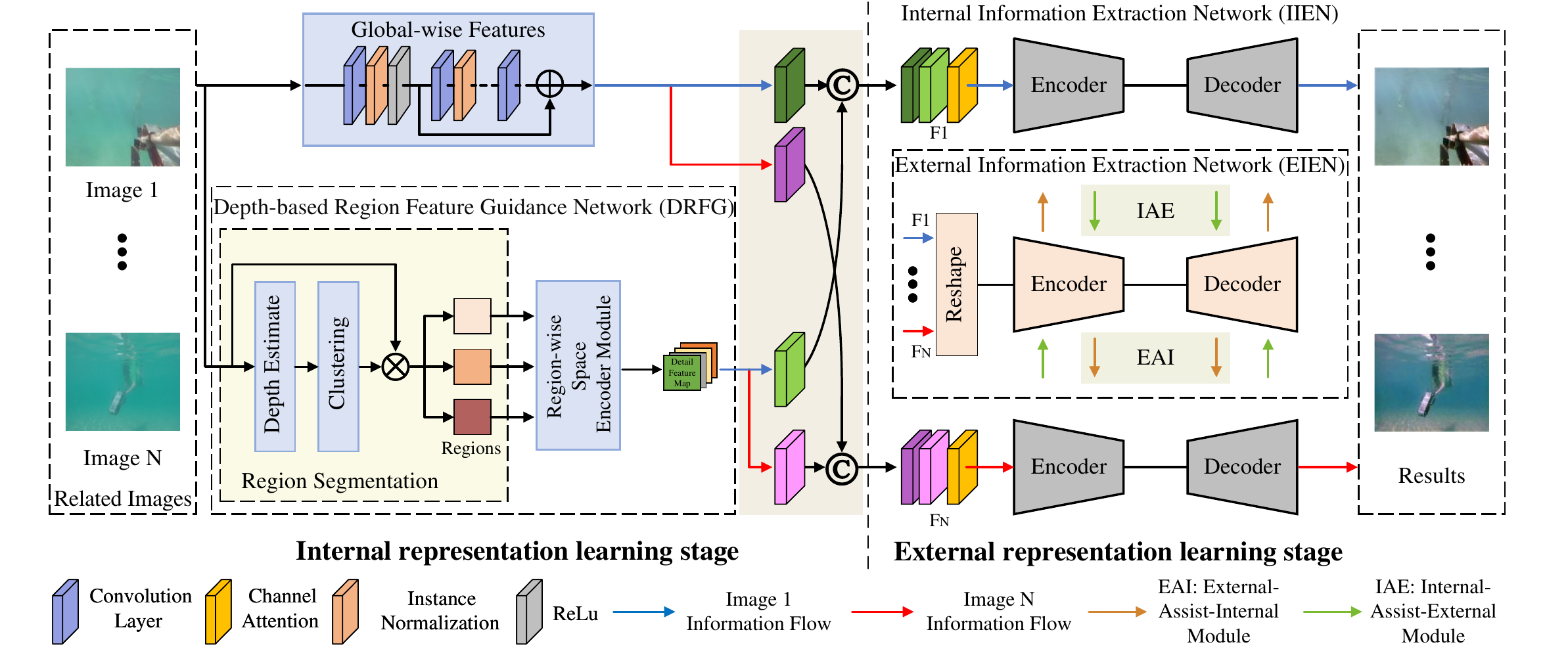}}
	\caption{\textbf{Flowchart of the proposed UIERL}, which contains an internal representation learning stage and an external representation learning stage. In the internal representation learning stage, a new depth-based region feature guidance network is designed to guide intra-image differentiated enhancement, which is an effective complement and guidance to global-wise features, including a region segmentation based on scene depth and a region-wise space encoder module. In the external representation learning stage, we first propose an external information extraction network to mine the rich external information in the related images. Then, two interaction modules are proposed, where internal and external features interact with each other via the proposed internal-assist-external module and external-assist-internal module. In this way, our UIERL incorporates rich internal and external information to better enhance a single image. }
	\label{Fig3}
\end{figure*}
 
\subsection{Internal-External Combined Learning}
To our best knowledge, there exists no internal-external combined learning work for underwater images. The closest area is multi-frame video super resolution. In the early years, extensive works~\cite{Dong2014, Kim2016, Lim2017, Lai2017} focus on single-frame internal information learning and propose many effective networks to improve the performance of super resolution. For example, Dong~\etal~\cite{Dong2014} propose SRCNN, which is the pioneering work of CNN-based single super resolution. Later, numerous variants of CNNs are widely explored, such as VDSR~\cite{Kim2016}, SAN~\cite{Zhao2020} and EDSR~\cite{Lim2017}. Recently, inspired by the finding that different observations of a same object or scene are highly likely to exist in consecutive frames of video, many multi-frame super-resolution methods (such as VSRnet~\cite{Li2017}, BRCN~\cite{Kappeler2016} and ESPCN~\cite{Shi2016}) are proposed. These methods further introduce the information provided by neighboring frames, \ie, external information, exceeding the limitation of single-frame methods and achieving state-of-the-art performance on video super-resolution.

Similarly, underwater images also have rich external information. Due to the uniqueness of the acquisition approach, underwater image acquisition tools usually capture multiple images in the same or similar scenes. Thus, the underwater images to be enhanced in practical usage are highly correlated. They can provide rich complementary information for each other to improve the enhancement performance. 

\section{Proposed Method}
\subsection{Overall Architecture}
The overview of the proposed UIERL is shown in Fig.\ref{Fig3}. The whole pipeline can be divided into two stages, \ie, the internal representation learning stage and the external representation learning stage. In the first stage, a new depth-based region feature guidance network is designed to provide rich differentiated enhancement guidance for global-wise features in the original input, including a region segmentation on scene depth and a region-wise space encoder module. Details are introduced in Section~\ref{section:internal}. In the second stage, an external information extraction network is developed to further mine the rich external information. After that, internal and external features interact with each other via the proposed external-assist-internal module and internal-assist-external module, obtaining richer features to better enhance a single image. Details are introduced in Section~\ref{section:external}.

\subsection{Internal Representation Learning Stage}
\label{section:internal}
\subsubsection{\textbf{Motivation}} Like most existing deep UIE models, we first learn the global-wise feature transformation for the whole image enhancement. However, an unavoidable concern is that only learning image-to-image enhancement from a global-wise view is unreasonable due to a significant region-wise quality difference within the underwater image. Regrettably, most existing methods neglect this phenomenon. In their results, regions with uneven degradation tend to be over-/under-enhanced, and even affect the overall enhancement effect. It is important to consider the local region quality difference while pursuing good global-wise enhancement.

The classical underwater image formation model explicitly characterizes the underwater imaging process as follows:
\begin{equation}
I=J \cdot e^{-\beta d}+A \cdot\left(1-e^{-\beta d}\right)
\end{equation}
where $I \in \mathbb{R}^{3 \times H \times W}$ denotes the captured underwater image with height $H$ and width $W$, and $J \in \mathbb{R}^{3 \times H \times W}$ is the clean dewatered image. $d \in \mathbb{R}^{1 \times H \times W}$ denotes the distance between the camera and the object, \ie, scene depth. $\beta \in \mathbb{R}^{1 \times 1 \times 3}$ indicates the attenuation coefficient, which is determined by the water body, and $A \in \mathbb{R}^{1 \times 1 \times 3}$ refers to the global ambient light. For a captured underwater image, the global ambient light and attenuation coefficient are essentially the same at all pixel locations within the image. Therefore, from the model, one could find that the region-wise quality difference within one underwater image is sensitive to the scene depth. In other words, the degradation degree of underwater images can be implicitly reflected by the scene depth~\cite{chiang2011underwater, Akkaynak2017, Akkaynak2018, Akkaynak2019}.

\begin{figure*}[!t]
	\centering
	\centerline{\includegraphics[width=0.99\linewidth]{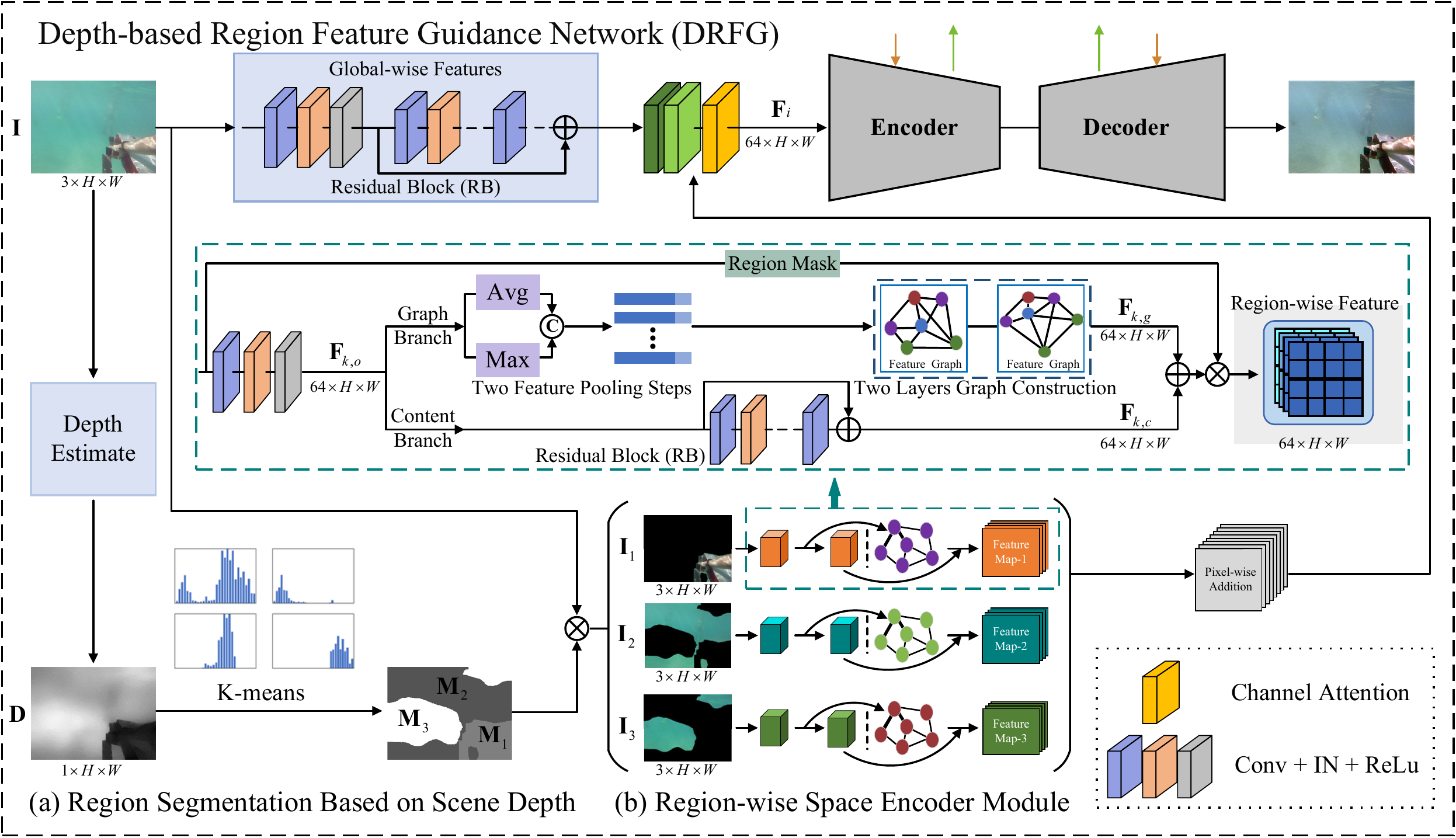}}
	\caption{\textbf{Overview of our DRFG proposed in the internal representation learning stage.} DRFG focuses on region-wise feature learning and provides an effective enhancement guidance for global-wise features to guide intra-image differentiated enhancement, which contains two components: (a) a region segmentation based on scene depth and (b) a region-wise space encoder module.}
	\label{Fig4}
\end{figure*}

Combining the above two points, a new depth-based region feature guidance network (DRFG) is proposed to guide intra-image differentiated enhancement, which can be an effective complement and guidance for global-wise enhancement. The pipeline of DRFG is shown in Fig.\ref{Fig4}, which includes two key components, \ie, the region segmentation based on scene depth and the region-wise space encoder module. First, the original input image is divided into $K$ non-overlapping regions by region segmentation based on scene depth, each with a similar quality degradation pattern. Then, regions with different quality are learned separately. Finally, by spatially combining all local region features, the whole region-wise guidance features for the input image can be obtained.

\subsubsection{\textbf{Region Segmentation}} In this part, benefiting from the fact that the degradation degree of underwater images is highly correlated with scene depth, the region segmentation operation is performed directly on the scene depth map. As shown in Fig.\ref{Fig4} (a), the depth map $\boldsymbol{D} \in \mathbb{R}^{1 \times H \times W}$ of the input underwater image $\boldsymbol{I} \in \mathbb{R}^{3 \times H \times W}$ is first estimated using the depth estimation network proposed in our previous work~\cite{Lin9311705}. Then, the obtained depth map is clustered at the pixel level using the K-means~\cite{LIKAS2003451} algorithm and classified into three region masks, denoted as $\boldsymbol{M}_{k} \in \mathbb{R}^{1 \times H \times W}$, $k=1,2,3$, representing three different quality degradation patterns. By multiplying the elements between the input image and the region segmentation masks, multiple regions $\boldsymbol{I}_{k} \in \mathbb{R}^{3 \times H \times W}$, $k=1,2,3$ of the input image are obtained, where each region has a similar scene depth values inside.

\subsubsection{\textbf{Region-wise Space Encoder Module}} As shown in Fig.\ref{Fig4} (b), this module considers the region-wise quality difference within the underwater image in two aspects. (1) Regions with different scene depth values usually have different degradation patterns and they should be enhanced separately. Therefore, the module contains three encoder paths that encode information from each of the three regions. In each encoder path, region-wise feature learning is performed within their own local context. (2) A region with similar scene depth values usually shares a similar quality degradation pattern. Such correlation inspires us that the information within the same region can cooperate with each other to extract useful region information. Therefore, a parallel content branch and a graph branch are embedded in each encoder path. The graph branch aims at exploiting the quality degradation similarity within the same region in the feature space, and combines with the content branch to obtain discriminative local region features.
\begin{figure*}[!t]
	\centering
	\centerline{\includegraphics[width=0.99\linewidth]{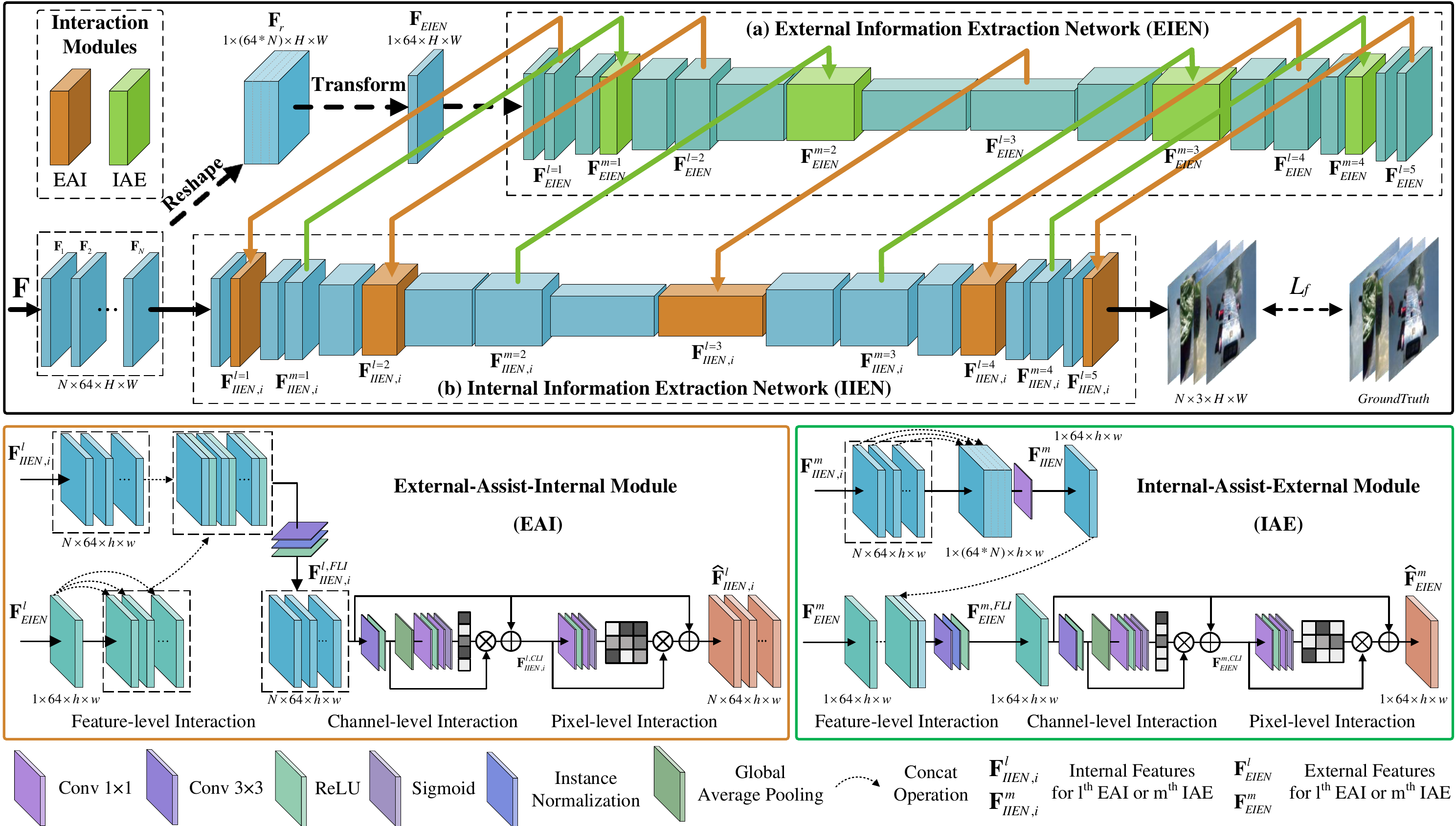}}
	\caption{\textbf{Overview of the proposed external representation learning stage}, which consists of two branches with mutual interaction. The top branch, called external information extraction network (EIEN), is used to mine the rich external information in the related images. The bottom branch, called internal information extraction network (IIEN), is used to mine the internal information of a single image. The two branches interact with each other via the proposed external-assist-internal module (EAI) and internal-assist-external (IAE) module. In this way, the proposed UIERL incorporates rich internal and external information to better enhance a singe image.}
	\label{Fig5}
\end{figure*}

Taking one encoder path as an example, the initial feature of the region input $\boldsymbol{I}_{k}$ is extracted by a stacked Conv-IN-ReLU unit. Then, the extracted initial region features $\boldsymbol{F}_{k, o}$ are imported into the content branch and graph branch, respectively. The content branch consists of a residual block with two stacked Conv-IN-ReLU units and a residual connection to further extract regional content features $\boldsymbol{F}_{k, c} \in \mathbb{R}^{64 \times H \times W}$, which can be expressed as
\begin{equation}
\boldsymbol{F}_{k, c}=\mathcal{C}_{cont}\left(\boldsymbol{F}_{k, o})\right),
\end{equation}
where $\mathcal{C}_{cont}$ denotes the processing of the content branch.

The graph branch builds a dense two-layer region graph to take all input initial region features as the node representations. Meanwhile, each edge of the graph constructs the interactions between any pair-wise nodes regardless of their positional distance, thereby well capturing more dependencies. During network optimization, each node is updated based on both its neighbors and itself, from which the rich cooperative information in the same degradation region can be fully mined and combined, improving the consistency of the enhancement of regions with the same quality degradation attribute. To reduce the memory and computational costs brought by the nodes, a down-sampling part, including Max pooling and Average Pooling operations, is performed on the initial region features $\boldsymbol{F}_{k, o}$ to retain different important features and avoid redundancy as much as possible. Then, a new region features $\boldsymbol{F}_{k, g}$ will be acquired as follows
\begin{equation}
\begin{aligned}
\boldsymbol{F}_{k, g} & =\mathcal{C}_{graph}\left(\boldsymbol{F}_{k, o}\right) \\
& =\mathcal{F}_{softmax}\left(\hat{\mathbf{A}}^{k} \operatorname{ReLU}\left(\hat{\mathbf{A}}^{k} \boldsymbol{F}_{k, o} \mathbf{W}_{1}^{k}\right) \mathbf{W}_{2}^{k}\right),
\end{aligned}
\end{equation}
where $\mathcal{C}_{graph}$ denotes the processing of the graph branch. $\mathbf{W}_{1}^{k} \in \mathbb{R}^{d^{k} \times c_{1}^{k}}$, $\mathbf{W}_{2}^{k} \in \mathbb{R}^{c_{1}^{k} \times c^{k}}$ refer to the learnable weight matrices of two fully-connected layers for feature projections. $\hat{\mathbf{A}}^{k}=\tilde{\mathbf{D}^{k}}^{-\frac{1}{2}} \tilde{\mathbf{A}}^{k} \tilde{\mathbf{D}^{k}}^{-\frac{1}{2}}$, where $\tilde{\mathbf{D}}^{k}(i, i)=\sum_{j} \tilde{\mathbf{A}}^{k}(i, j)$ is the degree matrix of $\tilde{\mathbf{A}}^{k}$ that is diagonal. $\tilde{\mathbf{A}}^{k}=\mathbf{A}^{k}+\mathbf{I}^{k}$, where $\mathbf{I}^{k}$ is the identity matrix, $\mathbf{A}^{k}$ is the learnable adjacency matrix. Inspired by adaptive graph learning techniques and self-attention mechanisms, for the constructed region graph, to learn a task-specific graph structure, the learnable adjacency matrix is defined as follows 
\begin{equation}\label{eq4}
{\mathbf{A}^{k}=\sigma \left(\boldsymbol{F}_{k, o} \mathbf{P}_{1}^{k}\left(\boldsymbol{F}_{k, o} \mathbf{P}_{2}^{n}\right)^{\top}\right)}
\end{equation}
where $\mathbf{P}_{1}^{k}$ and $\mathbf{P}_{2}^{k}$ are two learnable matrices used to reduce the computational cost, and $\sigma $ is the sigmoid function.

Finally, the output local region features of the encoder path are obtained by adding the feature representations of both the content and the graph branches. Then, the local region features of three sub-regions are combined spatially, and then fed back to the global-wise features extracted from the original resolution. These region features have obvious differences and advantages, which can be used as a complementary tool to guide differentiated enhancement for uneven quality-degraded regions within the image.

\subsection{External Representation Learning Stage} 
\label{section:external}
\subsubsection{\textbf{Motivation}} Unlike diverse image acquisition methods on land, underwater image acquisition tools usually capture multiple images in the same or similar scenes. In practical applications, the underwater images to be enhanced are highly correlated. Consequently, it is possible to take advantage of the rich external information in the related images to enhance the quality of a single image.

To this end, an external representation learning stage is proposed, which includes two branches with mutual interaction, as shown in Fig.\ref{Fig5}. The top branch, termed as external information extraction network (EIEN), aims to extract features from multiple related images, \ie, external information. The bottom branch, termed as internal information extraction network (IIEN), aims to extract features from a single image, \ie, internal information. Both top and bottom branches are designed based on the commonly used dense-Unet network. Additionally, a reshape operation, a transform operation and four internal-assist-external modules (IAE) are added to the top branch, and five external-assist-internal modules (EAI) are added to the bottom branch. The two branches interact with each other via the proposed EAI and IAE. 

Let $\boldsymbol{I}=\left\{I_1, I_2 \cdots I_N\right\} \in \mathbb{R}^{N \times 3 \times H \times W}$ be a given batch of images with width $W$ and height $H$. $N$ is the number of images in the batch. Taking advantage of the internal representation learning stage, the discriminative single image representation is generated, denoted as $\boldsymbol{F}=\left\{F_1, F_2, ..., F_N\right\} \in \mathbb{R}^{N \times 64 \times H \times W}$. In the top branch, these features are first concatenated along the channel dimension (\ie, the reshape operation in Fig.\ref{Fig5} (a)) and get $\boldsymbol{F}_{r} \in \mathbb{R}^{1 \times (64*N) \times H \times W}$. Then, the channel dimension $64*N$ of $\boldsymbol{F}_{r}$ is adaptively reduced into 64 (\ie, the transform operation in Fig.\ref{Fig5} (a)), and these features are further input into EIEN to extract multiple image features $\boldsymbol{F}_{EIEN} \in \mathbb{R}^{1 \times 64 \times H \times W}$. In the bottom branch, each image is input into IIEN to obtain single feature representations $\boldsymbol{F}_{IIEN, i} \in \mathbb{R}^{N \times 64 \times H \times W}, i=1,2 \cdots N$. Then, $\boldsymbol{F}^{l}_{EIEN}$ from the top branch interacts with $\boldsymbol{F}^{l}_{IIEN, i}$ by EAI. Moreover, $\boldsymbol{F}^{m}_{IIEN, i}$ is utilized to update $\boldsymbol{F}^{m}_{EIEN}$ by IAE. In this way, the features of the two branches are enhanced by jointly exploring the internal-external information interaction. In the following, we give details for the proposed EAI and IAE.
 
\subsubsection{\textbf{Proposed External-Assist-Internal Module (EAI)}} As shown in Fig.\ref{Fig5} (a), the top branch aims at extracting the rich external feature by EIEN. In the bottom branch, the $l$-th EAI employs rich external feature  $\boldsymbol{F}^{l}_{EIEN} \in \mathbb{R}^{1 \times 64 \times h \times w}, h \in[0, H], w \in[0, W]$  to assist each single image feature representation $\boldsymbol{F}^{l}_{IIEN, i} \in \mathbb{R}^{N \times 64 \times h \times w}, i=1,2 \cdots N$, as shown in the orange box. EAI contains three steps, \ie, feature-level interaction, channel-level interaction and pixel-level interaction. Specifically, $\boldsymbol{F}^{l}_{EIEN}$ and each $\boldsymbol{F}^{l}_{IIEN, i}$ are concatenated along the channel dimension, and a stacked Conv-IN-ReLU unit is utilized to fuse each concatenated feature, 
\begin{equation}
\boldsymbol{F}^{l,FLI}_{IIEN,i}=\mathcal{C}_{conv} \left(Concat\left(\boldsymbol{F}^{l}_{EIEN}, \boldsymbol{F}^{l}_{IIEN, i}\right)\right)
\end{equation}
where $\boldsymbol{F}^{l,FLI}_{IIEN,i}$ is generated by the independent feature interaction with $\boldsymbol{F}^{l}_{EIEN}$; $\mathcal{C}_{conv}$ denotes the stacked Conv-IN-ReLU unit, and $Concat (\cdot)$ means the concatenation operation.

In addition to exploring the feature-level interaction, we also conduct interactions along the channel and spatial dimensions. After obtaining the output feature $\boldsymbol{F}^{l,FLI}_{IIEN,i}$ of the feature-level interaction, the channel-level interaction first uses a stacked Conv-ReLU unit to generate the preliminary feature, and then embeds a global average pooling (G) to produce channel-wise summary statistics. After that, two convolution layers (C), a ReLU (R) and a Sigmoid ($\sigma$) operation are performed to capture the channel-wise interaction dependencies. By adding the residual back to the raw input feature, the feature enhanced by the channel-level interaction can be formulated as:
\begin{equation}
\begin{aligned}
& \boldsymbol{W}^{l,CLI}_{IIEN,i} = \sigma \left(C\left(R\left(C\left(G\left(\boldsymbol{F}^{l,FLI}_{IIEN,i}\right)\right)\right)\right)\right. \\
& \boldsymbol{F}^{l,CLI}_{IIEN,i} = \boldsymbol{W}^{l,CLI}_{IIEN,i} * \boldsymbol{F}^{l,FLI}_{IIEN,i} + \boldsymbol{F}^{l, FLI}_{IIEN,i}
\end{aligned}
\end{equation}

Similar to the channel-level interaction, the pixel-level interaction takes feature $\boldsymbol{F}^{l,CLI}_{IIEN,i}$ as input, and the interaction weight of each pixel is calculated by two convolution layers (C) with ReLU (R) and Sigmoid ($\sigma$) operations. By adding the residual back to the raw input feature, the feature enhanced by the pixel-level interaction can be formulated as:
\begin{equation}
\begin{aligned}
& \boldsymbol{W}^{l,PLI}_{IIEN,i} =\sigma \left(C\left(R\left(C\left(\boldsymbol{F}^{l,CLI}_{IIEN,i})\right)\right)\right)\right. \\
&\hat{\boldsymbol{F}}^{l}_{IIEN,i} = \boldsymbol{W}^{l,PLI}_{IIEN,i} * \boldsymbol{F}^{l,CLI}_{IIEN,i} + \boldsymbol{F}^{l,FLI}_{IIEN,i}
\end{aligned}
\end{equation}
where $*$ represents multiplication operation, and $\hat{\boldsymbol{F}}^{l}_{IIEN,i} (i=1, 2 \cdots, N)$ is the final output of feature, channel and pixel-level interactions with the help of $\boldsymbol{F}^{l}_{EIEN}$.

\subsubsection{\textbf{Proposed Internal-Assist-External Module (IAE)}} As shown in Fig.\ref{Fig5} (b), the bottom branch extracts single image feature representations by IIEN. In the top branch, the $m$-th IAE employs single image representation feature $\boldsymbol{F}^{m}_{IIEN, i} \in \mathbb{R}^{N \times 64 \times h \times w}, i=1,2 \cdots N$  to assist related image feature representation $\boldsymbol{F}^{m}_{EIEN} \in \mathbb{R}^{1 \times 64 \times h \times w}, h \in[0, H], w \in[0, W]$, as shown in the green box. Similar to EAI, IAE also contains three steps, \ie, feature-level interaction, channel-level interaction and pixel-level interaction. First, a reshape operation (\ie, channel-wise concatenation) is applied on $\boldsymbol{F}^{m}_{IIEN, i} \in \mathbb{R}^{N \times 64 \times h \times w}, i=1, 2 \cdots N$ to generate the concatenated feature $\boldsymbol{F}^{m}_{IIEN} \in \mathbb{R}^{1 \times (64*N) \times h \times w}$. Then $\boldsymbol{F}^{m}_{IIEN}$ is fed to a 1x1 convolution layer to reduce the dimension, \ie, $\boldsymbol{F}^{m}_{IIEN} \in \mathbb{R}^{1 \times 64 \times h \times w}$. After that, $\boldsymbol{F}^{m}_{IIEN}$ and $\boldsymbol{F}^{m}_{EIEN}$ are concatenated along the channel dimension, and a stacked Conv-IN-ReLU unit is added to process the concatenated feature,
\begin{equation}
\boldsymbol{F}^{m,FLI}_{EIEN}=\mathcal{C}_{conv} \left(Concat\left(\boldsymbol{F}^{m}_{EIEN}, \boldsymbol{F}^{m}_{IIEN}\right)\right)
\end{equation}
where $\boldsymbol{F}^{m,FLI}_{EIEN}$ is updated by the feature interaction with $\boldsymbol{F}^{m}_{EIEN}$; $\mathcal{C}_{conv}$ denotes the stacked Conv-IN-ReLU unit, and $Concat (\cdot)$ means the concatenation operation. Similar to EAI, we also perform channel and pixel-level interactions on top of the features generated by the feature-level interaction, and represent the output features of IAE as $\hat{\boldsymbol{F}}^{m}_{EIEN}$.

In summary, we make use of the two interaction modules to incorporate rich internal and external information, thereby better enhancing a single image.
 
\subsection{Loss Function}
Our proposed UIERL is an end-to-end network, where the content and perceptual loss are combined to achieve both good quantitative and qualitative scores. Given a batch of training pairs $\left\{\boldsymbol{I}_n, \boldsymbol{J}_n \right\}_{n=1}^N$ that contain $N$ related images and their corresponding clean counterparts, $L_{1}$ loss is first adopted as the content loss to reduce the pixel difference between the enhancement result and the corresponding ground truth, which can well restore the sharpness of edges and details:
\begin{equation}
L_{1}=\frac{1}{N} \sum_{n=1}^N\left\|\text {UIERL}\left(\boldsymbol{I}_n \right)-\boldsymbol{J}_n \right\|_1
\end{equation}

Besides, the perceptual loss $L_{per}$ is used to reduce the feature difference between the enhanced result and the ground truth image, which is computed based on the VGG-19 network pre-trained on ImageNet:
\begin{equation}
L_{per}=\frac{1}{N} \sum_{n=1}^N (\|\phi \left(\text {UIERL}\left(\boldsymbol{I}_n \right)\right)-\phi \left(\boldsymbol{J}_n\right)\|_2)
\end{equation}
where $\phi(\cdot)$ represents the feature maps of the pool-3 layer of the pre-trained VGG-19 network. Finally, the final loss function can be expressed as:
\begin{equation}\label{eq13}
L_{f}=\lambda_{1} * L_{1} + \lambda_{2} * L_{per}
\end{equation}
where $\lambda_{1}$ and $\lambda_{2}$ are trade-off weights, and they are empirically set to 0.8 and 0.2 to balance different losses.

\section{Experiments}
In this section, the implementation details and experiment settings are first described. Then, quantitative and qualitative analyses are performed to evaluate the effectiveness of the proposed UIERL. Finally, a series of ablation studies are provided to verify the advantages of each component.

\subsection{Implementation Details}
As described above, existing underwater datasets for testing do not fully reflect the characteristics of underwater test data in practical applications since they usually contain only one test image per scene, as shown in Fig.\ref{Fig2} (b). To this end, a large-scale real underwater image scene dataset (UISD) is developed, and each scene contains multiple images. To be specific, a large number of underwater video scene clips are first collected from YouTube, Google and our self-captured videos, and then carefully selected and refined. Most of the collected video scenes are weeded out, and about 1513 scenes of 6,064 images are retained, including different image content, color ranges and degrees of contrast decrease. For each underwater scene, the number of captured underwater images is different, ranging from 2 to 17. 

To train our UIERL, we first randomly select 313 scenes of 1,571 images from the proposed UISD dataset, and construct corresponding enhancement references in a manual selection manner similar to the UIEB dataset~\cite{li2019underwater}. Then, 313 scenes are randomly divided into training and test pairs. In detail, 253 underwater scenes containing 1300 images are used for training, denoted as~\textbf{Train-R1300}, and 60 underwater scenes containing 271 images for testing, denoted as~\textbf{Test-R271}. Both in the training and testing phase, all inputs are resized to 256x256 and the pixel values are normalized between [-1, 1]. To further expand the training samples and avoid overfitting, random horizontal flipping and rotation are performed as data augmentation in the training phase.

Our UIERL is an end-to-end framework. The entire framework is implemented on the PyTorch platform, and all experiments are conducted on an Nvidia Quadro RTX 8000 Ti GPU (48G memory). Adam with a learning rate of 1e-4 is used to optimize the network, and the default values of $\beta_{1}$ and $\beta_{2}$ are set to 0.5 and 0.999, respectively. For each training iteration of the training stage, we do not limit the number of a batch, the images in each batch are all randomly selected from a same or similar scene. In the testing stage, each image scene with an arbitrary quantity of related images constitutes a batch.

\subsection{Experiment Settings}
\emph{\textbf{1) Benchmarks.}} To verify the generalization ability of the proposed UIERL on real-world underwater images in practical applications, five real underwater image datasets, including \textbf{SQUID}, \textbf{sub-RUIE}, \textbf{sub-UIEB}, \textbf{sub-EUVP} and a more comprehensive real-world underwater scene dataset \textbf{UISD-1200} are validated.
\begin{itemize}
\item The SQUID~\cite{Berman2020UnderwaterSI} dataset includes 57 underwater stereo image pairs taken from four different dive sites in Israel, \ie, 114 images. These images are divided into 35 scenes, and each scene has 2-10 images. 
\item The sub-RUIE/sub-UIEB/sub-EUVP is the image subset that is more consistent with the characteristics of underwater test data in practical applications, which is selected from the most commonly used real underwater image datasets RUIE\cite{liu2019real} / UIEB\cite{li2019underwater} / EUVP\cite{islam2020fast}. The sub-RUIE dataset includes 413 scenes of 1524 images, and each scene has 2-20 images. The sub-UIEB dataset includes 112 scenes of 267 images, and each scene has 2-4 images. The sub-EUVP dataset includes 99 scenes of 358 images, and each scene has 2-5 images. 
\item The UISD-1200 dataset is the rest 1200 scenes of the proposed UISD dataset, which includes 4496 images and each scene has 2-17 images.
\end{itemize}

All the above datasets and the UIERL code are available at \emph{\url{ https://github.com/Underwater-Lab-SHU/UIERL}}.

\emph{\textbf{2) Compared Methods.}} To verify the performance of our method, 8 state-of-the-art (SOTA) deep learning-based UIE methods are compared, including WaterNet (TIP’19)~\cite{li2019underwater}, FUIEGAN (RAL’2020)~\cite{islam2020fast}, LCNet (TMM’2021)~\cite{Jiang2022a}, Ucolor (TIP’21)\cite{Li2021}, TOPAL (TCSVT’2022)\cite{Jiang2022}, CLUIE (TCSVT’2022)~\cite{Li2022}, PUIE-MC (ECCV’2022)~\cite{Fu2022} and SGUIE (TIP’2022)~\cite{Qi2022}. Note that for all the above-mentioned methods, the publicly released test models and parameters are used to produce their results.

\emph{\textbf{3) Evaluation Metrics.}} To quantitatively evaluate the enhancement performance, four no-reference underwater quality assessment metrics, including CCF, UIQM, UCIQE and Edge Intensity (Edge), are used. A higher CCF or UIQM or UCIQE score suggests a better human visual perception. A higher Edge score denotes a better edge intensity. To more accurately evaluate the performance, a user study is conducted to measure the perceptual scores (PS) of different methods. 100 images are first randomly selected from each of the above test datasets. Then, 10 volunteers are invited to score the enhancement results of each method under the same environment. The score range is set from 1 to 5, representing ”Bad”, ”Poor”, ”Fair”, ”Good” and ”Excellent”, respectively.

Since color deviation is an important characteristic of underwater images, we also evaluate the color restoration accuracy on the 16 representative examples presented in the project page of SQUID~\cite{Berman2020UnderwaterSI}, denoted as SQUID-16. The SQUID-16 dataset includes four dive sites (Nachsholim, Michmoret, Katzaa and Satil, denoted as Set A, Set B, Set C and Set D, respectively), where four representative samples are selected from each dive site. The authors also provide the accurate color parameters of the corresponding color card, and calculate the average angle reproduction error for color recovery evaluation. The smaller the color error, the better the color correction performance.
\begin{table*}[!htbp]
	\centering
	\small
	\caption{CCF/UCIQE/UIQM/Edge Intensity (Edge)/Perceptual Scores (PS) of different methods on four real underwater test datasets. }
	\renewcommand\arraystretch{1.2}
	\setlength\tabcolsep{6.2pt}
	\begin{tabular}{C{0.07\linewidth}lcccccccccc}
		\Xhline{1.2pt}
		\rule{0pt}{10pt}
		Dataset & Metric &
		\makecell[c]{Inputs}&
		\makecell[c]{WaterNet}& 
		\makecell[c]{FUIE-GAN}& 
		\makecell[c]{LCNet}& 
		\makecell[c]{Ucolor}& 
		\makecell[c]{TOPAL}&
		\makecell[c]{CLUIE}&
		\makecell[c]{PUIE-MC}&
		\makecell[c]{SGUIE}&
		\makecell[c]{Ours}
		\\
		\hline
		\multirow{5}*{SQUID} 
		& Edge $\uparrow$ &
		18.164	&29.777	&33.351	&{\textcolor{blue}{34.564}}	&31.359	&28.732	&29.601	&31.220	&33.035	&\textbf{\textcolor{red}{49.076}}
		\\
		& UIQM $\uparrow$ &
		-1.0574	&2.0363	&1.4822	&{\textcolor{blue}{2.8812}}	&1.9124	&1.7969	&1.2513	&2.6109	&1.2127	&\textbf{\textcolor{red}{3.9014}}
		\\
		& CCF $\uparrow$ &
		12.827&	17.573&	18.808&	20.880&	20.170&	19.213&	19.173&	20.179&	{\textcolor{blue}{21.676}}&	\textbf{\textcolor{red}{29.771}}
		\\
		& UCIQE $\uparrow$ &
		0.4054&	0.4943&	0.4883&	0.4981&	0.5012&	0.4788&	0.4720&	0.5129&	{\textcolor{blue}{0.5376}}&	\textbf{\textcolor{red}{0.5817}}
		\\
		& PS $\uparrow$ &
		1.3740&	2.8120&	2.2940&	2.0200&	2.8290&	2.4040&	2.7590&	\textcolor{blue}{2.8520}&	2.6990&	\textbf{\textcolor{red}{4.1440}}

		\\
		\hline
		\multirow{5}*{sub-RUIE} 
		& Edge $\uparrow$ &
		39.887	&61.310	&61.472	&65.184	&60.861	&64.452	&73.581	&66.822	&{\textcolor{blue}{77.828}}	&\textbf{\textcolor{red}{86.058}}
		\\
		& UIQM $\uparrow$ &
		2.3924	&4.5477	&{\textcolor{blue}{4.9809}}	&4.8638	&4.2518&	4.4049	&4.4980	&4.5239	&3.8876	&\textbf{\textcolor{red}{5.1218}}
		\\
		& CCF $\uparrow$ &
		18.850&	28.020&	26.948&	28.922&	30.034&	29.968&	33.864&	32.016&	{\textcolor{blue}{34.470}}&	\textbf{\textcolor{red}{38.718}}
		\\
		& UCIQE $\uparrow$ &
		0.4495&	0.5443&	0.5239&	0.5200&	0.5361&	0.5143&	0.5476&	0.5455&
		\textbf{\textcolor{red}{0.5817}}&	{\textcolor{blue}{0.5611}}
		\\
		& PS $\uparrow$ &
		1.4200&	2.7990&	2.5400&	2.1400&	2.8300&	2.7540&	2.7890&	\textcolor{blue}{3.0860}&	2.4810&	\textbf{\textcolor{red}{4.1350}}

		\\
		\hline
		\multirow{5}*{sub-UIEB}
		& Edge $\uparrow$ &
		42.490&	61.681&	62.192&	68.444&	69.936&	70.297&	{\textcolor{blue}{82.553}}&	71.059&	81.823&	\textbf{\textcolor{red}{86.404}}
		\\
		& UIQM $\uparrow$ &
		3.0823&	4.5024&	{\textcolor{blue}{4.8017}}&	4.4348&	4.5559&	4.3390&	4.5217&	4.4181&	4.1000&\textbf{\textcolor{red}{4.9794}}	
		\\
		& CCF $\uparrow$ &
		20.289	&27.520	&27.215&	29.938	&32.849	&32.275	&{\textcolor{blue}{36.710}}	&33.049	&35.989	&\textbf{\textcolor{red}{39.084}}
		\\
		& UCIQE $\uparrow$ &
		0.4745	&0.5589	&0.5454	&0.5319&	0.5713	&0.5521&	0.5759&	0.5696&\textbf{\textcolor{red}{0.6017}}		&{\textcolor{blue}{0.5966}}
		\\
		& PS $\uparrow$ &
		1.7230&	2.4960&	2.3450&	2.1420&	2.8720&	2.6810&	\textcolor{blue}{2.8880}&	2.8370&	2.7000&	\textbf{\textcolor{red}{3.7280}}
		
		\\
		\hline
		\multirow{5}*{sub-EUVP}
		& Edge $\uparrow$ &
		25.597&	50.889&	50.388&	44.062&	48.483&	42.400&	52.433&	48.616&{\textcolor{blue}{60.505}}	&	\textbf{\textcolor{red}{73.545}}
		\\
		& UIQM $\uparrow$ &
		1.4322&	3.4550&	{\textcolor{blue}{4.0575}}&	3.4529&	3.1306&	3.0022&	3.1146&	3.4919&	2.6176&	\textbf{\textcolor{red}{4.7130}}
		\\
		& CCF $\uparrow$ &
		11.954&	22.823&	20.430&	19.536&	22.215&	19.461&	23.518&	22.486&{\textcolor{blue}{27.519}}&	\textbf{\textcolor{red}{33.131}}
		\\
		& UCIQE $\uparrow$ &
		0.3871&	0.5455&	0.4912&	0.4532&	0.5104&	0.4595&	0.4928&	0.5132&	{\textcolor{blue}{0.5721}}	&	\textbf{\textcolor{red}{0.5857}}
		\\
		& PS $\uparrow$ &
		1.3740&	2.8120&	2.2940&	2.020&	2.8290&	2.4040&	2.7590&	\textcolor{blue}{2.8520}&	2.6990&	\textbf{\textcolor{red}{4.1440}}
		
		\\
		\Xhline{1.2pt}
	\end{tabular}
	\label{table1} 
	\begin{tablenotes}
		\footnotesize
		\item The \textbf{\textcolor{red}{best}} and \textcolor{blue}{second-best} results are marked in red and blue respectively.
	\end{tablenotes}
\end{table*}
\begin{table*}[!htbp]
	\centering
	\small
	\caption{CCF/UCIQE/UIQM/Edge Intensity (Edge)/Perceptual Scores (PS) of different methods on the proposed UISD-1200 dataset. }
	\renewcommand\arraystretch{1.2}
	\setlength\tabcolsep{6.37pt}
	\begin{tabular}{C{0.07\linewidth}lcccccccccc}
		\Xhline{1.2pt}
		\rule{0pt}{13pt}
		Dataset & Metric &
		\makecell[c]{Inputs}&
		\makecell[c]{WaterNet}& 
		\makecell[c]{FUIE-GAN}& 
		\makecell[c]{LCNet}& 
		\makecell[c]{Ucolor}& 
		\makecell[c]{TOPAL}&
		\makecell[c]{CLUIE}&
		\makecell[c]{PUIE-MC}&
		\makecell[c]{SGUIE}&
		\makecell[c]{Ours}
		\\
		\hline
		\multirow{5}*{UISD-1200} 
		& Edge $\uparrow$ &
		49.019&	54.890&	61.513&	65.159&	54.382&	67.108&	\textcolor{blue}{73.097}&	63.408&	68.502&	\textbf{\textcolor{red}{74.897}}	
		\\
		& UIQM $\uparrow$ &
		3.2450&	4.0749&	\textcolor{blue}{4.3551}&	3.6810&	4.1234&	3.9448&	4.0735&	3.9082&	3.7635&	\textbf{\textcolor{red}{4.4810}}
		\\
		& CCF $\uparrow$ &
		24.630&	25.601&	28.281&	30.214&	27.492&	32.589&	\textcolor{blue}{35.151}&	31.254&	33.582&	\textbf{\textcolor{red}{36.321}}			
		\\
		& UCIQE $\uparrow$ &
		0.5338&	0.5531&	0.5580&	0.5504&	0.5570&	0.5824&	0.5939&	0.5792&	\textbf{\textcolor{red}{0.6079}}&	\textcolor{blue}{0.5961}	
		\\
		& PS $\uparrow$ &
		2.0650&	2.3370&	2.5400&	2.3760&	2.5560&	2.7520&	2.8020&	\textcolor{blue}{2.8290}&	2.7610&	\textbf{\textcolor{red}{3.7930}}
				
		\\
		\Xhline{1.2pt}
	\end{tabular}
	\begin{tablenotes}
		\footnotesize
		\item The \textbf{\textcolor{red}{best}} and \textcolor{blue}{second-best} results are marked in red and blue respectively.
	\end{tablenotes}
	\label{table2} 
\end{table*}

\subsection{Comparison With SOTA UIE Methods}
In this subsection, we conduct quantitative and qualitative comparisons on diverse test datasets to evaluate the effectiveness of the proposed UIERL. Besides, the accuracy of color restoration is analyzed. Due to the limited space, more results are given in the supplementary material.

\textbf{Quantitative Comparisons.} The quantitative results of different methods on real challenging sets are reported in Table~\ref{table1} and Table~\ref{table2}. As reported, our method outperforms the compared models in terms of most metrics on five datasets. For example, CCF, UIQM and Edge scores of the proposed model consistently outperform all compared models. For the UCIQE scores, the proposed method achieves the best performance on the sub-EUVP and SQUID datasets, and is only inferior to SGUIE on the sub-UIEB, sub-RUIE and UISD-1200 datasets, ranking second. For the most challenging dataset UISD-1200, the proposed method still shows strong competitive performance compared with other models. Observing the perceptual scores of different methods, the proposed method obtains the highest perceptual score on five datasets, and is obviously superior to the competing methods. Such results also further suggest the visually pleasing quality of our results. Other deep methods achieve similar perceptual scores. Among them, Ucolor, CLUIE, PUIE-MC and SGUIE perform relatively well due to introducing underwater prior information. However, they do not fully explore the significant characteristics of region-wise quality differences within the image, and thus their performance is limited. In contrast, our method effectively exploits the internal characteristic of underwater images, and further combines the rich external information provided by related images, achieving a significant gain.

\begin{figure*}[!t]
	\centering
	\centerline{\includegraphics[scale=0.500]{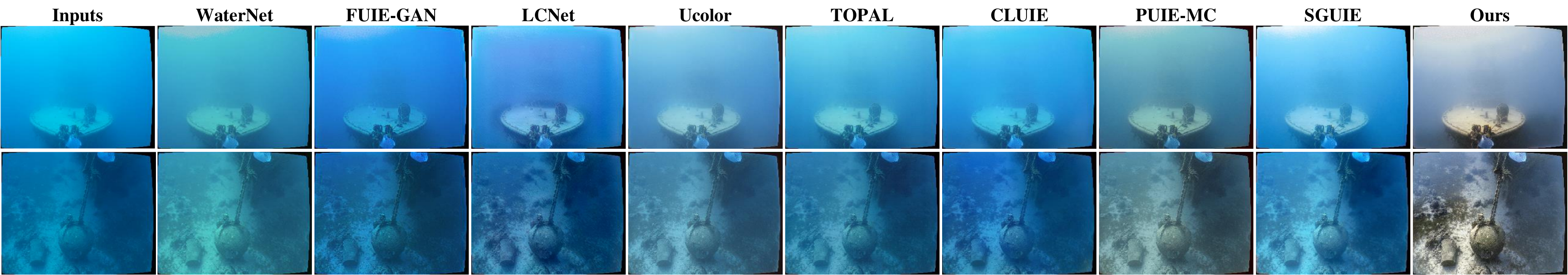}} 
	\caption{Visual comparisons on challenging underwater images sampled from \textbf{SQUID}. From left to right are raw underwater images, and the results of WaterNet~\cite{li2019underwater}, FUIEGAN~\cite{islam2020fast}, LCNet~\cite{Jiang2022a}, Ucolor~\cite{Li2021}, TOPAL~\cite{Jiang2022}, CLUIE~\cite{Li2022}, PUIE-MC~\cite{Fu2022}, SGUIE~\cite{Qi2022} and our proposed UIERL.}
	\label{Fig_SQUID} 
\end{figure*}
\begin{figure*}[!t]
	\centering
	\centerline{\includegraphics[scale=0.500]{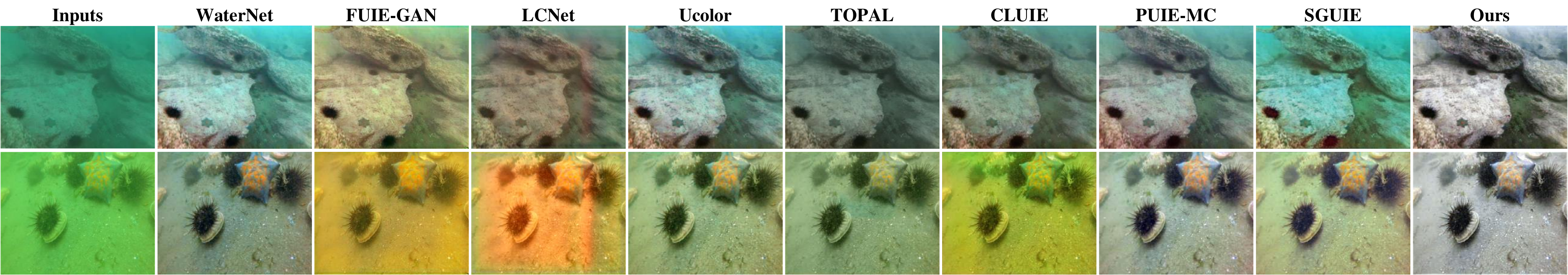}} 
	\caption{Visual comparisons on challenging underwater images sampled from \textbf{sub-RUIE}. From left to right are raw underwater images, and the results of WaterNet~\cite{li2019underwater}, FUIEGAN~\cite{islam2020fast}, LCNet~\cite{Jiang2022a}, Ucolor~\cite{Li2021}, TOPAL~\cite{Jiang2022}, CLUIE~\cite{Li2022}, PUIE-MC~\cite{Fu2022}, SGUIE~\cite{Qi2022} and our proposed UIERL.}
	\label{Fig_sub-RUIE} 
\end{figure*}
\begin{figure*}[!t]
	\centering
	\centerline{\includegraphics[scale=0.500]{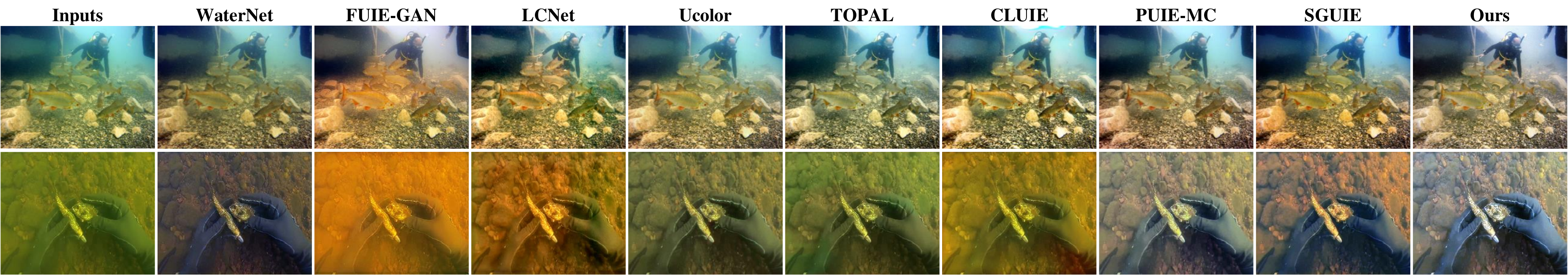}} 
	\caption{Visual comparisons on challenging underwater images sampled from \textbf{sub-UIEB}. From left to right are raw underwater images, and the results of WaterNet~\cite{li2019underwater}, FUIEGAN~\cite{islam2020fast}, LCNet~\cite{Jiang2022a}, Ucolor~\cite{Li2021}, TOPAL~\cite{Jiang2022}, CLUIE~\cite{Li2022}, PUIE-MC~\cite{Fu2022}, SGUIE~\cite{Qi2022} and our proposed UIERL.}
	\label{Fig_sub-UIEB} 
\end{figure*}
\begin{figure*}[!t]
	\centering
	\centerline{\includegraphics[scale=0.500]{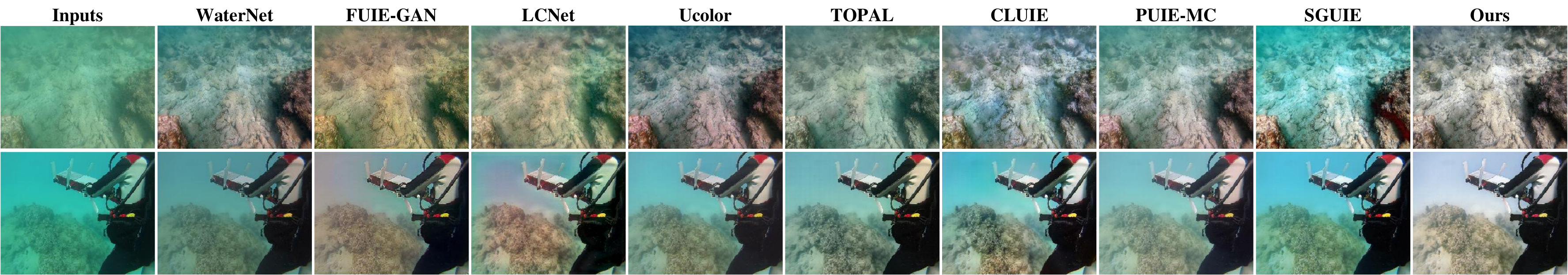}} 
	\caption{Visual comparisons on challenging underwater images sampled from \textbf{sub-EUVP}. From left to right are raw underwater images, and the results of WaterNet~\cite{li2019underwater}, FUIEGAN~\cite{islam2020fast}, LCNet~\cite{Jiang2022a}, Ucolor~\cite{Li2021}, TOPAL~\cite{Jiang2022}, CLUIE~\cite{Li2022}, PUIE-MC~\cite{Fu2022}, SGUIE~\cite{Qi2022} and our proposed UIERL.} 
	\label{Fig_sub-EUVP} 
\end{figure*}
\begin{figure*}[!t]
	\centering
	\centerline{\includegraphics[scale=0.500]{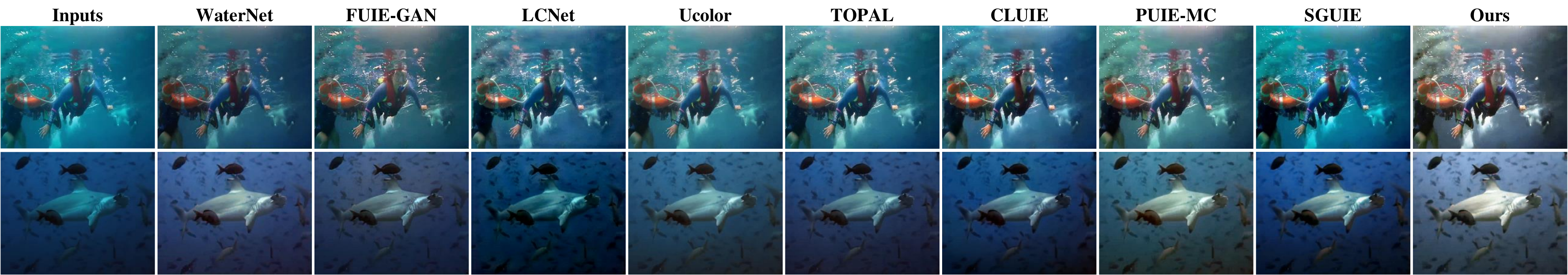}} 
	\caption{Visual comparisons on challenging underwater images sampled from \textbf{UISD-1200}. From left to right are raw underwater images, and the results of WaterNet~\cite{li2019underwater}, FUIEGAN~\cite{islam2020fast}, LCNet~\cite{Jiang2022a}, Ucolor~\cite{Li2021}, TOPAL~\cite{Jiang2022}, CLUIE~\cite{Li2022}, PUIE-MC~\cite{Fu2022}, SGUIE~\cite{Qi2022} and our proposed UIERL.}
	\label{Fig_UISD-1200} 
\end{figure*}

\textbf{Qualitative Comparisons.}~We also perform some visual comparisons with other methods on underwater images sampled from five benchmarks, and the results are shown in Fig.\ref{Fig_SQUID}-Fig.\ref{Fig_UISD-1200}. These images cover a wide range of underwater image characteristics (\eg, color casts, low contrast and blurred details), including blueish images, greenish images, yellowish images, blue-green and some complex images. Results of the SQUID and sub-RUIE datasets are first shown in Fig.\ref{Fig_SQUID} and Fig.\ref{Fig_sub-RUIE}. For these underwater images with obvious blueish or greenish color deviation and heavy haze, the proposed method not only corrects the intrinsic color appearance, but also enhances the texture detail and contrast. All the methods under comparison either under-enhance the image or have some haze remains. In terms of color, Water-Net, Ucolor, PUIE-MC and SGUIE cannot restore the realistic color, and still exist severe color distortion in the foreground part. FUIE-GAN, LCNet and CLUIE even introduce extra color casts in the results. TOPAL enhances the images in a gloomy tone, which seems unrealistic in the real world. Besides, the competing methods cannot recover the complete scene structure, for example, the content details of the distant reef are not well restored. In contrast, our method can generate more pleasing results and clear details, which is credited to the effective designs of the internal and external representation learning stages.

Results of the \textbf{sub-UIEB} and \textbf{sub-EUVP} datasets are shown in Fig.\ref{Fig_sub-UIEB} and Fig.\ref{Fig_sub-EUVP}. For these images with diverse blue-green or yellowish color tones, the proposed method still shows notable superiority in both color correction and detail recovery. Other deep methods cannot obtain satisfactory results. Almost all comparison methods cannot handle color casts and blurriness well, even producing unreal colors and additional noise in the results, such as WaterNet, FUIE-GAN and LCNet. PUIE-MC corrects the color deviation to some extent, but the image contrast is not well improved. SGUIE improves the contrast of the input images but introduces obvious over-saturation, thereby resulting in limited visual effects.
\begin{table}[!t]
	\centering
	\small
	\caption{THE AVERAGE ANGULAR REPRODUCTION ERROR (AE) ON SQUID-16. TOP TWO RESULTS ARE IN RED AND BLUE. }
	\setlength\tabcolsep{6.6pt}
	\begin{tabular}{l|ccccc}
		\Xhline{1pt}
		\rule{0pt}{9pt}
		\multirow{2}{*}{Methods} & \multicolumn{5}{c}{\makecell[c]{Angular Reproduction Error (AE)}}
		\\
		\cline{2-6}
		\rule{0pt}{9pt}
		&Set A&Set B&Set C&Set D& Avge
		\\
		\hline
		\hline
		\rule{0pt}{9pt}
		WaterNet &21.112&21.547&19.703&23.198&21.390
		\\
		\rule{0pt}{9pt}
		FUIE-GAN &24.849&25.804&26.445&30.661&26.939
		\\
		\rule{0pt}{9pt}
		LCNet&\textbf{\textcolor{red}{6.884}}&11.620&12.151&17.863&\textcolor{blue}{12.129}
		\\
		\rule{0pt}{9pt}
		Ucolor&22.543&18.109&14.731&18.114&18.374
		\\
		\rule{0pt}{9pt}
		TOPAL&15.823&16.278&15.895&20.632&17.157
		\\
		\rule{0pt}{9pt}
		CLUIE&25.314&20.446&23.282&25.437&23.620
		\\
		\rule{0pt}{9pt}
		PUIE-MC&20.467&\textcolor{blue}{10.619}&\textcolor{blue}{11.581}&\textcolor{blue}{14.966}&14.408
		\\
		\rule{0pt}{9pt}
		SGUIE&30.947&29.606&24.522&25.965&27.760
		\\
		\hline
		\rule{0pt}{9pt}
		Our UIERL&\textcolor{blue}{10.567}&\textbf{\textcolor{red}{7.663}}&\textbf{\textcolor{red}{7.111}}&\textbf{\textcolor{red}{4.810}}&\textbf{\textcolor{red}{7.538}}
		\\
		\Xhline{1pt}
	\end{tabular}
	\label{SQUID-16_table} 
\end{table}
\begin{figure}[!t]
	\centering
	\hspace{-3.9pt}
	\centerline{\includegraphics[scale=0.55]{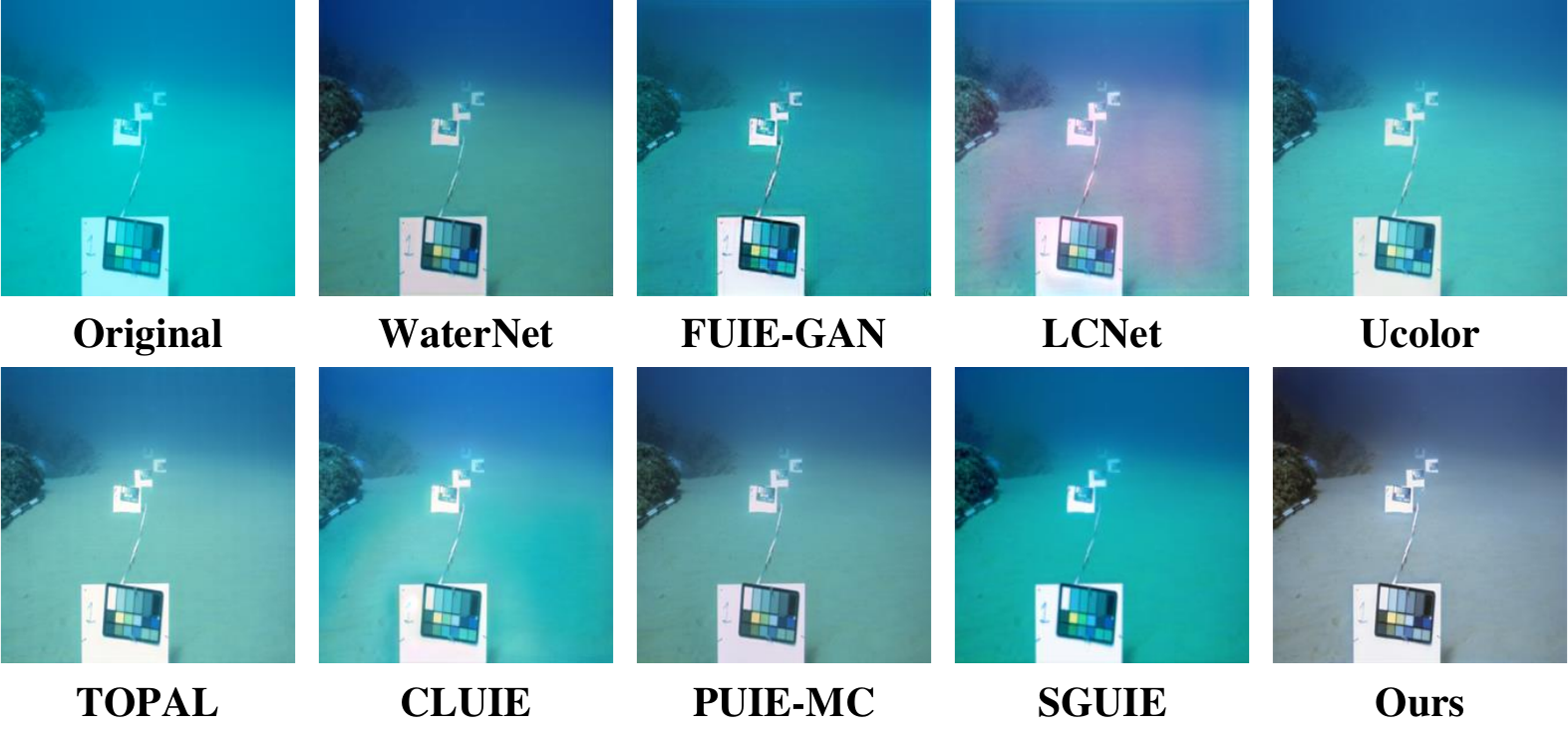}} 
	\caption{Visual comparisons on the \textbf{SQUID-16} dataset. Obviously, our results have a pleasing visual perception and good color restoration accuracy.}
	\label{Fig_SQUID-16} 
\end{figure}

Results of the \textbf{UISD-1200} dataset are shown in Fig.\ref{Fig_UISD-1200}. For these images captured on various challenging scenes, our method still can effectively correct colors, improve contrast and enhance details. Specifically, benefiting from the designed internal representation learning stage, our enhancement highlights the differential enhancement of different degraded regions, which makes the enhanced image overall look more natural (see the foreground and background regions of the first and second samples). Coupled with the rich external information in the related images, the proposed method can restore more details of the original images, such as the structure and edges of the object (see the human in the first sample and the shark in the second sample). In contrast, the competing methods show a limited effect on color correction and contrast enhancement. Ucolor, LCNet and PUIE-MC even introduce unrealistic colors in the results. CLUIE's performance is relatively good, but it cannot remove the color cast of the foreground, and fails to restore clear structure. All quantitative and qualitative results suggest that the proposed method can produce visually pleasing results and have more robust performance in different underwater scenes.

\textbf{Color Restoration Comparisons.}~Table~\ref{SQUID-16_table} reports the average color angular reproduction error on the SQUID-16 dataset. In the Set A subset, LCNet obtains the lowest color error while our UIERL ranks the second best. For the Set B, Set C and Set D subsets, the proposed method achieves the lowest average error. Besides, the proposed method obtains the best average performance across 16 images. Such results demonstrate the effectiveness of our method for underwater color correction. Fig.~\ref{Fig_SQUID-16} shows the comparisons of different methods on underwater images sampled from SQUID-16. As shown, our method not only provides pleasing visuals but also achieves quite good color restoration.

\subsection{Ablation Studies and Analysis}
In this subsection, a series of ablation studies are conducted to analyze the contribution of each proposed component. Compared with the SQUID, sub-UIEB and sub-EUVP datasets, the images of sub-RUIE (1524 images) and UISD-1200 (4496 images) datasets cover more diverse underwater scenes, which are more in line with realistic scenarios. Thus, we perform the ablation studies on sub-RUIE and UISD-1200 datasets.

\textbf{\emph{1) Effectiveness of the Internal and External Representation Learning Stage.}} The proposed UIERL includes two stages, \ie, the internal representation learning stage and the external representation learning stage. To verify the roles of these two stages, three ablation models are designed: 1) \textbf{M0}: without the internal and external representation learning stage; 2) \textbf{M1}: using the internal representation learning stage; 3) \textbf{M2}: using the internal and external representation learning stage, \ie, the whole UIERL. Results are reported in Table~\ref{table5}. It is clear that on both datasets, \textbf{M1} achieves significant gains in all metrics compared to \textbf{M0}. Such results demonstrate the effectiveness of enhancing regions with different quality in a divide-and-conquer manner in the internal representation learning stage. Observing the scores of \textbf{M2} and \textbf{M1}, we can see that the whole UIERL network achieves better performance on almost all metrics, with only a slight drop in the UIQM metric on the UISD-1200 dataset. This is because the external representation learning stage further exploits the rich external information provided by the related images and can better enhance a single image. 

\begin{figure}[!t]
	\centering
	\hspace{-6pt}
	{\includegraphics[width=0.976\linewidth]{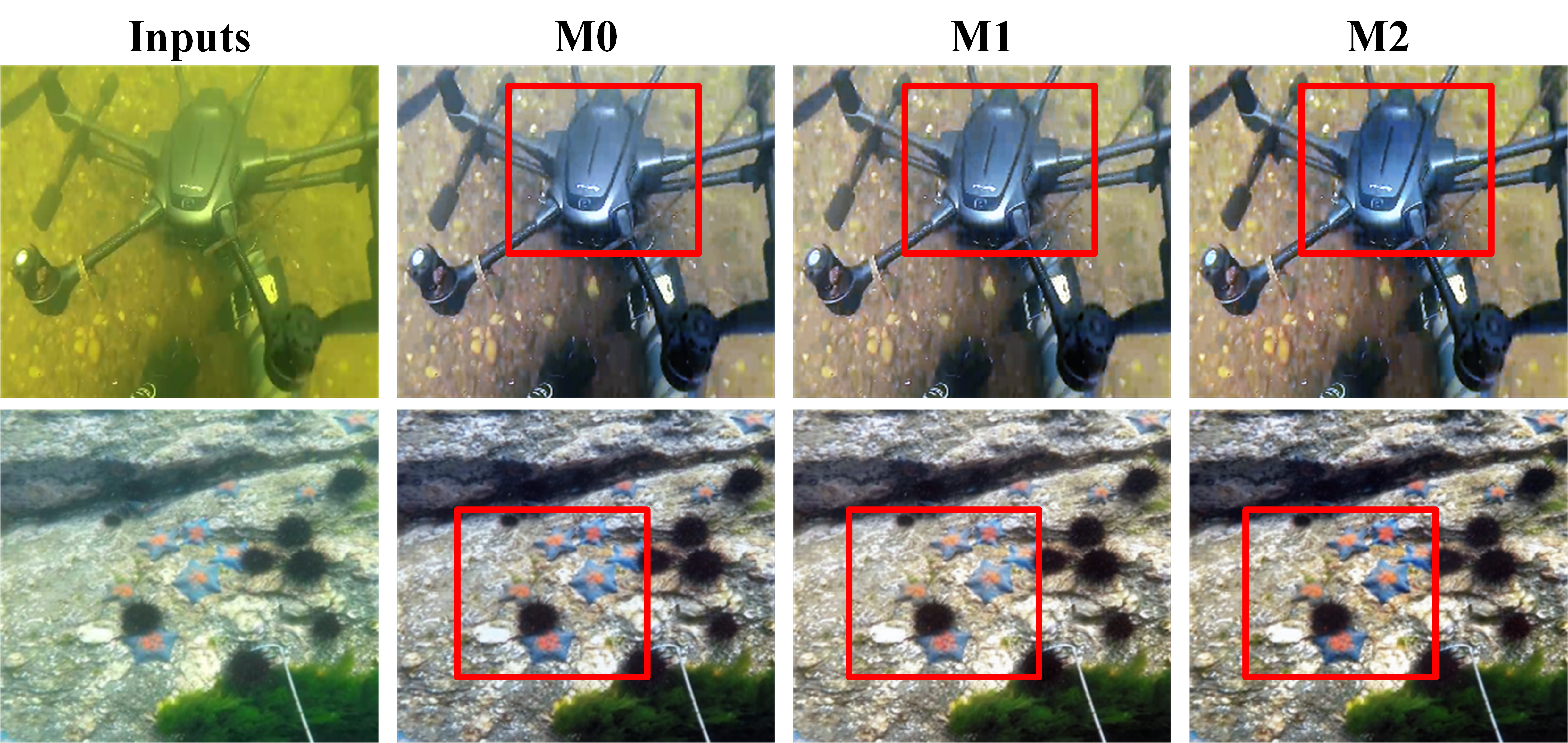}}
	\caption{{Visual examples of ablation study on the internal representation learning stage and the external representation learning stage.}}
	\label{Fig12}
\end{figure}

A few samples are shown in Fig.\ref{Fig12}. It can be seen that our internal representation learning stage can achieve differentiated enhancement for regions with different quality, such as foreground/background regions. Observing the fourth column, we can clearly see that by introducing rich external information to help a single image, the result is more vivid and natural in appearance and richer in details.
\begin{table}[!t]
	\centering
	\small
	\caption{ABLATION STUDY OF THE INTERNAL REPRESENTATION LEARNING STAGE AND EXTERNAL REPRESENTATION LEARNING STAGE. }
	\renewcommand\arraystretch{1.3}
	\setlength\tabcolsep{1.86pt}
	\begin{tabular}{l|cccc|cccc}
		\Xhline{1pt}
		\rule{0pt}{10pt}
		\multirow{2}{*}{Models}& \multicolumn{4}{c|}{sub-RUIE} & \multicolumn{4}{c}{UISD-1200}\\
		\cline{2-9}
		\rule{0pt}{10pt}
		& Edge & UIQM & CCF & UCIQE & Edge & UIQM & CCF & UCIQE\\
		\hline
		\rule{0pt}{10pt}
		M0&80.22&4.932&37.11&0.553&72.46&4.436&35.44&0.588\\
		\rule{0pt}{10pt}
		M1&\textcolor{blue}{84.08}&\textcolor{blue}{5.102}&\textcolor{blue}{38.05}&\textcolor{blue}{0.558}&\textcolor{blue}{74.58}&\textbf{\textcolor{red}{4.482}}&\textcolor{blue}{35.98}&\textcolor{blue}{0.591}\\
		\rule{0pt}{10pt}
		M2&\textbf{\textcolor{red}{86.05}}&\textbf{\textcolor{red}{5.121}}&\textbf{\textcolor{red}{38.71}}&\textbf{\textcolor{red}{0.561}}&\textbf{\textcolor{red}{74.89}}&\textcolor{blue}{4.481}&\textbf{\textcolor{red}{36.32}}&\textbf{\textcolor{red}{0.596}}\\
		\Xhline{1pt}
	\end{tabular}
	\label{table5} 
\end{table}
\begin{table}[!t]
	\centering
	\small
	\caption{ABLATION STUDY OF THE INTERNAL REPRESENTATION LEARNING STAGE. }
	\renewcommand\arraystretch{1.3}
	\setlength\tabcolsep{1.86pt}
	\begin{tabular}{l|cccc|cccc}
		\Xhline{1pt}
		\rule{0pt}{10pt}
		\multirow{2}{*}{Models}& \multicolumn{4}{c|}{sub-RUIE} & \multicolumn{4}{c}{UISD-1200}\\
		\cline{2-9}
		\rule{0pt}{10pt}
		& Edge & UIQM & CCF & UCIQE & Edge & UIQM & CCF & UCIQE\\
		\hline
		\rule{0pt}{10pt}
		M3&81.60&	5.061&	\textcolor{blue}{36.83}&	\textcolor{blue}{0.557}&	71.90&	4.452&	35.22&	\textcolor{blue}{0.592}\\
		\rule{0pt}{10pt}
		M4&80.71&	5.058&	36.60&	0.551&	\textbf{\textcolor{red}{75.14}}&	\textcolor{blue}{4.453}&	\textcolor{blue}{35.93}&	\textbf{\textcolor{red}{0.596}}\\
		\rule{0pt}{10pt}
		M5&\textcolor{blue}{83.55}&\textcolor{blue}{5.058}&36.56&0.553&72.04&4.412&34.79&0.589\\
		\rule{0pt}{10pt}
		M1&\textbf{\textcolor{red}{84.08}}&\textbf{\textcolor{red}{5.102}}&\textbf{\textcolor{red}{38.05}}&\textbf{\textcolor{red}{0.558}}&\textcolor{blue}{74.58}&\textbf{\textcolor{red}{4.482}}&\textbf{\textcolor{red}{35.98}}&0.591\\
		\Xhline{1pt}
	\end{tabular}
	\label{table6} 
\end{table}
\begin{table}[!t]
	\centering
	\small
	\caption{ABLATION STUDY OF THE INTERACTION STRATEGY OF THE EXTERNAL REPRESENTATION LEARNING STAGE. }
	\renewcommand\arraystretch{1.3}
	\setlength\tabcolsep{1.86pt}
	\begin{tabular}{l|cccc|cccc}
		\Xhline{1pt}
		\rule{0pt}{10pt}
		\multirow{2}{*}{Models}& \multicolumn{4}{c|}{sub-RUIE} & \multicolumn{4}{c}{UISD-1200}\\
		\cline{2-9}
		\rule{0pt}{10pt}
		& Edge & UIQM & CCF & UCIQE & Edge & UIQM & CCF & UCIQE\\
		\hline 
		\rule{0pt}{10pt}
		M6&80.27&5.036&36.69&0.551&\textcolor{blue}{74.20}&4.459&35.97&0.590\\
		\rule{0pt}{10pt}
		M7&78.93&	5.038&	36.62&	0.549&	72.83&	4.454&	35.66&	0.588\\
		\rule{0pt}{10pt}
		M8&\textcolor{blue}{83.28}&	\textbf{\textcolor{red}{5.139}}&	\textcolor{blue}{37.50}&	\textcolor{blue}{0.556}&	74.19&	\textbf{\textcolor{red}{4.492}}&	\textcolor{blue}{35.99}&	\textcolor{blue}{0.592}\\
		\rule{0pt}{10pt}
		M2&\textbf{\textcolor{red}{86.05}}&\textcolor{blue}{5.121}&\textbf{\textcolor{red}{38.71}}&\textbf{\textcolor{red}{0.561}}&\textbf{\textcolor{red}{74.89}}&\textcolor{blue}{4.481}&\textbf{\textcolor{red}{36.32}}&\textbf{\textcolor{red}{0.596}}\\
		\Xhline{1pt}
	\end{tabular}
	\label{table7} 
\end{table}

\textbf{\emph{2) Effectiveness of Different Components of the Internal Representation Learning Stage.}} In the internal representation learning stage, DRFG is a key component to guide intra-image differentiated enhancement, which is achieved by a region segmentation based on scene depth and a region-wise space encoder module. In the region-wise space encoder module, a graph branch is embedded into each encoder path to extract discriminative local region feature representation within their own local context. To verify the effectiveness of these components, we perform the following ablation settings: 1) \textbf{M1}: using the internal representation learning stage with full DRFG; 2) \textbf{M3}: based on \textbf{M1}, removing the region segmentation based on the scene depth map and replacing it with the region segmentation based on the input original image; 3) \textbf{M4}: based on \textbf{M1}, removing the region-wise operation in the region-wise space encoder module, and using the same encoder module for feature extraction in all regions; 4) \textbf{M5}: based on \textbf{M1}, removing the graph branch in each encoder path in the region space encoder module, and using the content branch for local region feature extraction. Results are reported in Table \ref{table6}. Compared with \textbf{M1}, \textbf{M3} suffers from performance degradation on both sub-RUIE and UISD-1200 datasets. The results indicate that the region segmentation based on scene depth map can better perceive different quality regions than the segmentation based on the original image. The performance of \textbf{M4} is also lower when compared to \textbf{M1}, which indicates that differentiated processing of different quality regions is effective within one single image. Observing the scores of \textbf{M1} and \textbf{M5}, it can be seen that introducing a well-designed graph branch in each encoder path for better region-wise feature learning is necessary.
\begin{table}[!t]
	\centering
	\small
	\caption{ABLATION STUDY OF THE DIFFERENT COMPONENTS OF THE EAI/IAE MODULE PROPOSED IN THE INTERACTION STRATEGY. }
	\renewcommand\arraystretch{1.3}
	\setlength\tabcolsep{1.86pt}
	\begin{tabular}{l|cccc|cccc}
		\Xhline{1pt}
		\rule{0pt}{10pt}
		\multirow{2}{*}{Models}& \multicolumn{4}{c|}{sub-RUIE} & \multicolumn{4}{c}{UISD-1200}\\
		\cline{2-9}
		\rule{0pt}{10pt}
		& Edge & UIQM & CCF & UCIQE & Edge & UIQM & CCF & UCIQE\\
		\hline 
		\rule{0pt}{10pt}
		M6&80.27&5.036&36.69&0.551&\textcolor{blue}{74.20}&4.459&\textcolor{blue}{35.97}&0.590\\
		\rule{0pt}{10pt}
		M9&\textcolor{blue}{82.08}&\textcolor{blue}{5.098}&\textcolor{blue}{37.55}&\textcolor{blue}{0.557}&74.00&\textcolor{blue}{4.460}&35.90&\textcolor{blue}{0.591}\\
		\rule{0pt}{10pt}
		M2&\textbf{\textcolor{red}{86.05}}&\textbf{\textcolor{red}{5.121}}&\textbf{\textcolor{red}{38.71}}&\textbf{\textcolor{red}{0.561}}&\textbf{\textcolor{red}{74.89}}&\textbf{\textcolor{red}{4.481}}&\textbf{\textcolor{red}{36.32}}&\textbf{\textcolor{red}{0.596}}\\
		\Xhline{1pt}
	\end{tabular}
	\label{table8} 
\end{table}

\textbf{\emph{3) Effectiveness of Different Components of the External Representation Learning Stage.}} In the external representation learning stage, the interaction strategy is a key component to explore and exploit the rich external information provided by the related images, which is achieved by the proposed EAI and IAE. To verify their effectiveness, we conduct the following ablation studies: 1) \textbf{M2}: using the complete interaction strategy, \ie, the whole UIERL; 2) \textbf{M6}: based on \textbf{M2}, removing the IAE and EAI, and replacing them with the concatenation operation. 3) \textbf{M7}: based on \textbf{M2}, removing the IAE and replacing it with the concatenation operation; 4) \textbf{M8}: based on \textbf{M2}, removing the EAI and replacing it with the concatenation operation. Results are reported in Table~\ref{table7}. Compared with \textbf{M6}, the full model \textbf{M2} achieves the best performance across two datasets, which proves that the proposed interaction strategy can efficiently introduce rich external information to improve the enhancement performance of a single image. Comparing the scores of \textbf{M2}, \textbf{M7} and \textbf{M8}, we can see that the performance of most metrics in both datasets decreases after removing the EAI and IAE modules, which implies the effectiveness of the combinations of the proposed EAI and IAE modules.

\textbf{\emph{4) Contributions of Different Components of the EAI/IAE module.}} To verify the roles of the different components in the EAI/IAE module, we conduct the following ablation studies: 1) \textbf{M2}: using the complete EAI/IAE module with FLI, CLI and PLI, \ie, the whole UIERL; 2) \textbf{M6}: using the EAI/IAE module with FLI only; 3) \textbf{M9}: using the EAI/IAE module with FLI and CLI only. Results are shown in Table~\ref{table8}. As shown, the quantitative performance of the full model \textbf{M2} is significantly better than the ablation models \textbf{M6} and \textbf{M9}, which further suggests the necessity of interacting in different ways.

\section{Conclusion}
In this paper, a novel internal-external representation learning network is proposed for enhancing underwater images, which consists of two stages, \ie, an internal representation learning stage and an external representation learning stage. Firstly, a new depth-based region feature guidance network is proposed in the first stage to perceive and enhance the response of the network towards different quality-degraded regions, building a region feature guidance for intra-image differentiated enhancement. In the second stage, considering the characteristics of underwater test data in real applications, when processing a single image, we introduce the rich external information provided by the related images. The internal and external information interact with each other via the proposed external-assist-internal module and internal-assist-external module. Extensive experiments on five real datasets further demonstrate the superiority of the proposed method. 

\bibliographystyle{ieeetr}
\bibliography{UIERL}
\end{document}